\documentclass{article} 
\usepackage{arxiv,times}

\usepackage{amsmath,amsfonts,bm}

\def\eqref#1{equation~\ref{#1}}

\def\1{\bm{1}}

\DeclareMathAlphabet{\mathsfit}{\encodingdefault}{\sfdefault}{m}{sl}
\SetMathAlphabet{\mathsfit}{bold}{\encodingdefault}{\sfdefault}{bx}{n}

\usepackage{hyperref}
\usepackage{url}
\usepackage{bm}
\usepackage{wrapfig}
\usepackage[shortlabels]{enumitem}
\usepackage{subcaption}
\usepackage{graphicx}
\usepackage{tcolorbox}
\usepackage[table]{xcolor}  
\usepackage{booktabs}
\definecolor{color_blue}{HTML}{E7EFFA}
\definecolor{color_green}{HTML}{E6F8E0}
\definecolor{color_gray}{HTML}{ECECEC}
\definecolor{pearDark}{HTML}{2980B9}

\title{TempFlow-GRPO: When Timing Matters for GRPO in Flow Models}

\author{%
Xiaoxuan He$^{1,2}$\thanks{{} {} Equal Contribution.}, \hspace{.3em}
Siming Fu$^{1}$\footnotemark[1], \hspace{.3em}
Yuke Zhao$^{1}$\footnotemark[1], \hspace{.3em}
Wanli Li$^{1}$, \hspace{.3em}
Jian Yang$^{2}$, \hspace{.3em}\\
\textbf{Dacheng Yin$^{2}$}\thanks{{} {} Project Leader.}, \hspace{.3em}
\textbf{Fengyun Rao$^{2}$}, \hspace{.3em} 
\textbf{Bo Zhang$^{1}$}\thanks{{} {} Corresponding authors.} \hspace{.3em} \\
[1ex]
$^{1}$ ZheJiang University, \\
$^{2}$ WeChat Vision, Tencent Inc \\
}

\iclrfinalcopy
\begin{document}
\maketitle
\lhead{Preprint.}
\begin{figure}[ht]
\vspace{-16pt}
\begin{center}
	\includegraphics[width=0.97\linewidth]{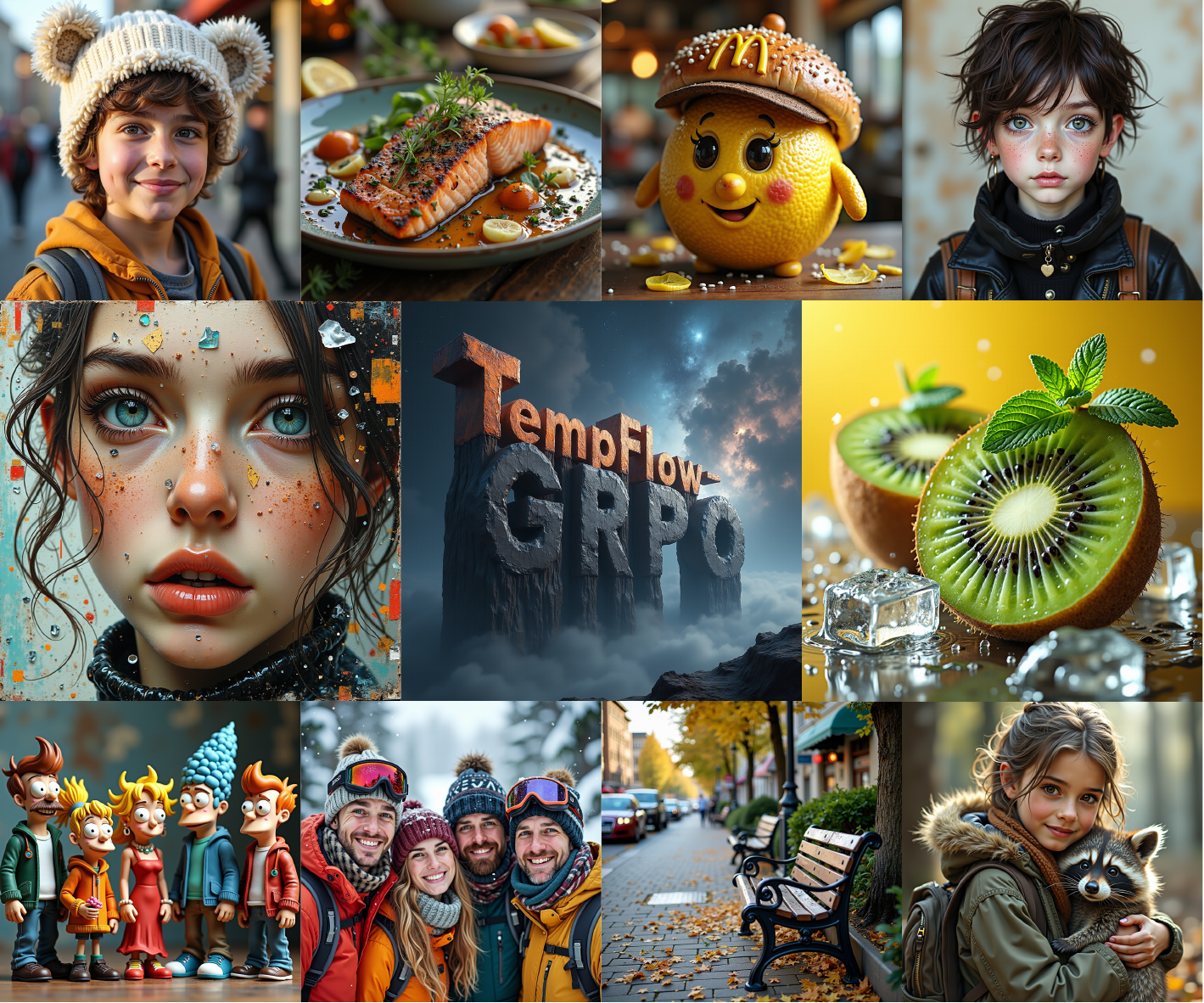}
\end{center}
\caption{Images generated by our proposed TempFlow-GRPO with FLUX.1-dev. It substantially improves the baseline models, achieving superior photorealism and enhanced fine-grained detail.}
\label{teaser}
\end{figure}


\begin{abstract}
Recent flow matching models for text-to-image generation have achieved remarkable quality, yet their integration with reinforcement learning for human preference alignment remains suboptimal, hindering fine-grained reward-based optimization. We observe that the key impediment to effective GRPO training of flow models is the temporal uniformity assumption in existing approaches: sparse terminal rewards with uniform credit assignment fail to capture the varying criticality of decisions across generation timesteps, resulting in inefficient exploration and suboptimal convergence. To remedy this shortcoming, we introduce \textbf{TempFlow-GRPO} (Temporal Flow-GRPO), a principled GRPO framework that captures and exploits the temporal structure inherent in flow-based generation. TempFlow-GRPO introduces three key innovations: (i) a trajectory branching mechanism that provides process rewards by concentrating stochasticity at designated branching points, enabling precise credit assignment without requiring specialized intermediate reward models; (ii) a noise-aware weighting scheme that modulates policy optimization according to the intrinsic exploration potential of each timestep, prioritizing learning during high-impact early stages while ensuring stable refinement in later phases; and (iii) a seed group strategy that controls for initialization effects to isolate exploration contributions. These innovations endow the model with temporally-aware optimization that respects the underlying generative dynamics, leading to state-of-the-art performance in human preference alignment and text-to-image benchmarks. Codes are available in \href{https://tempflowgrpo.github.io}{TempFlow-GRPO}.
\end{abstract}

\section{Introduction}

While text-to-image diffusion models have achieved unprecedented visual quality and semantic control~\citep{esser2024scaling, xie2024sana, labs2025flux}, aligning their outputs with human preference remains a formidable challenge. Reinforcement learning has emerged as a promising solution, giving rise to the field of Diffusion RL~\citep{wallace2024diffusion, black2023training, fan2023dpok}. However, the performance of these methods remains suboptimal, hindered by two fundamental limitations that have been largely overlooked: \textbf{\textit{ignoring the temporal dynamics of generation}} and \textbf{\textit{lacking intermediate feedback signals}}. These approaches apply uniform optimization across all timesteps and provide rewards only at completion, missing the varying importance of decisions throughout the generation process.

\begin{figure}[t]
    \centering
    \begin{subfigure}[b]{0.33\textwidth}
        \centering
        \includegraphics[width=\textwidth]{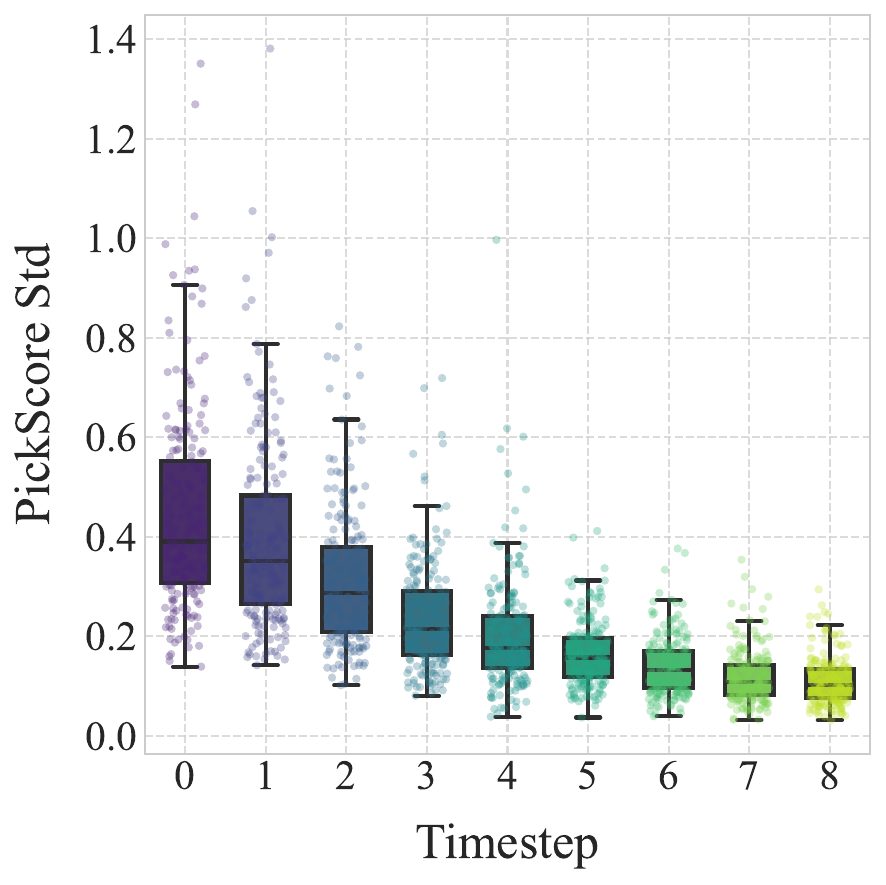}
        \label{fig1_1}
    \end{subfigure}
    \hfill
    \begin{subfigure}[b]{0.62\textwidth}
        \centering
        \includegraphics[width=\textwidth]{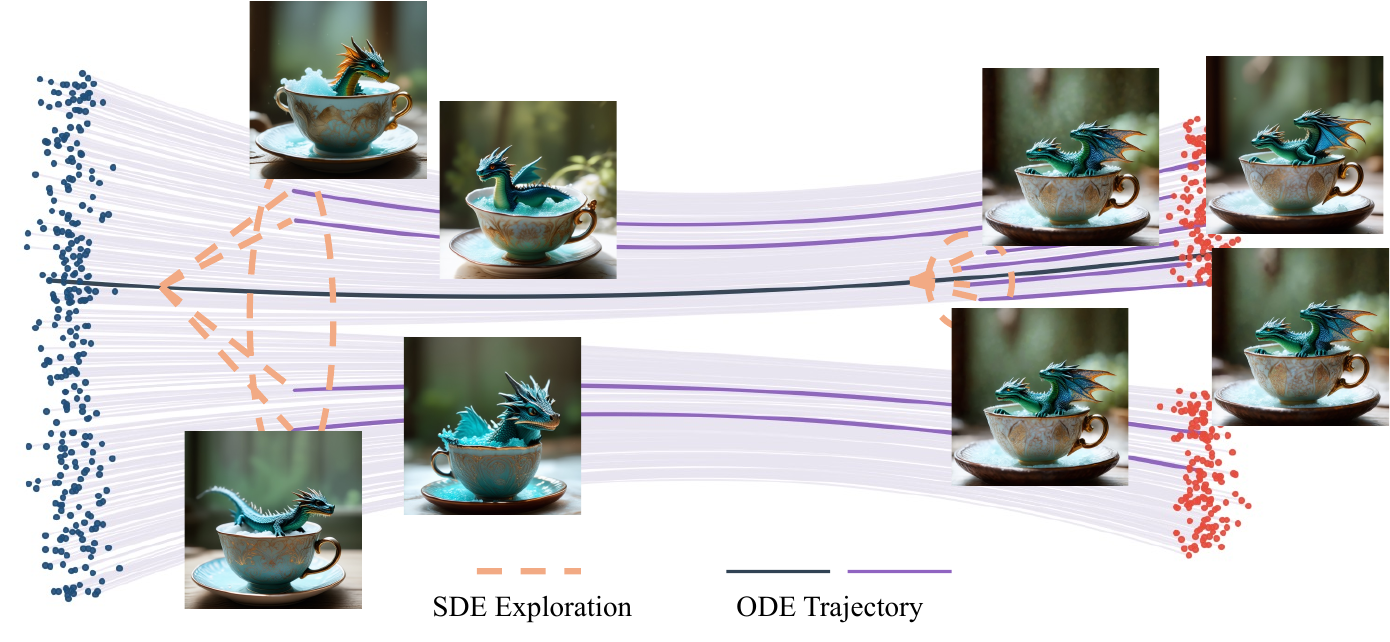}
        \label{fig1_2}
    \end{subfigure}
    \vspace{-0.5cm}
    \caption{(Left) Reward Variance Analysis: We plot the standard deviation of PickScore at each denoising step for 200 prompts, per prompt group size is 24. The results, obtained via applying SDE at only one step, reveal that reward variance is highest in the initial steps, indicating that early-stage interventions are most impactful for exploration. (Right) Method Illustration: By branching a stochastic (SDE) exploration from a specific, known state on a deterministic (ODE) trajectory, the resulting difference in the final reward can be unambiguously attributed to the exploration action taken at that precise branching point.}
    \label{fig1}
    \vspace{-0.5cm}
\end{figure}

The majority of existing approaches~\citep{gu2024diffusion, hong2024margin}, including recent works like Flow-GRPO~\citep{liu2025flow} and DanceGRPO~\citep{xue2025dancegrpo}, treat the multi-step generation process as a ``black box" with temporally agnostic optimization. They apply uniform updates across all timesteps despite the fact that each timestep operates under different noise conditions and contributes differently to final image quality. Specifically, we plot the Figure~\ref{fig1} (left) with applying SDE at only one timestep in the entire ODE trajectory, which ensures that any deviation in the final reward can be attributed to the stochastic exploration introduced at that specific step. As shown in Figure~\ref{fig1} (left), the std of the reward varies dramatically between timesteps, reaching a peak during early structural decisions (steps 0-2) and approaching zero during final refinements (steps 6-8). Yet Flow-GRPO maintains uniform treatment throughout, squandering high-impact exploration opportunities. Alternative approaches like SPO~\citep{liang2025aesthetic} attempt to address temporal dynamics through process reward models, but training such models on semantically ambiguous intermediate states is notoriously difficult. This raises a fundamental question: \textit{\textbf{how can we effectively achieve precise credit assignment for intermediate actions while adapting optimization intensity to each timestep's exploration capacity?}}

We address these limitations with \textbf{TempFlow-GRPO}, a temporally-aware RL framework built on two key insights. \textbf{First}, we define the visualization method in the left of Figure~\ref{fig1} as trajectory branching, which enables precise credit assignment by strategically introducing stochasticity at individual timesteps while maintaining deterministic evolution elsewhere (Figure~\ref{fig1}, Right). This provides \emph{provable} guarantees: (1) reward variance localizes to the branching point, (2) improvements are directly attributable to specific exploration outcomes, and (3) existing reward models require no modification. \textbf{Second}, noise-aware policy weighting modulates optimization intensity based on each timestep's intrinsic noise level. Early high-noise stages receive larger weight updates to encourage structural exploration, while late low-noise stages receive gentler updates to preserve learned features. \textbf{Third}, we introduce a seed-level grouping strategy that controls for initial noise influence, ensuring reward variations are attributed to branching exploration rather than random initialization. Together, these mechanisms create a framework that is conceptually simple, computationally efficient, and seamlessly integrates into existing flow matching architectures—all while respecting the temporal dynamics that uniform approaches ignore.

Our main contributions are threefold:

\begin{itemize}
    \item We pinpoint temporal uniformity—the equal treatment of all timesteps—as the primary limitation of flow-based GRPO. Our proposed \textbf{TempFlow-GRPO} overcomes this by introducing two key innovations: precise credit assignment to intermediate actions and noise-aware adaptation of optimization intensity.
    \item We introduce \textbf{trajectory branching} and \textbf{noise-aware reweighting} mechanisms to learn temporally-structured policies that respect the inherent dynamics of generative models. Additionally, we propose an efficient \textbf{seed group strategy} that effectively isolates exploration effects and considerably enhances overall performance.
    \item We demonstrate state-of-the-art performance on standard text-to-image benchmarks, achieving superior sample quality, human preference alignment, and compositional image generation compared to existing flow-based RL methods.
\end{itemize}

\section{Related Work}

\noindent \textbf{Alignment for Diffusion Models.} Alignment for diffusion models has emerged as a key research area. D3PO~\citep{yang2024using} introduces Direct Preference for Denoising Diffusion Policy Optimization to directly fine-tune diffusion models. Diffusion-DPO~\citep{wallace2024diffusion} adapts Direct Preference Optimization (DPO~\citep{rafailov2023direct}) as a simpler RLHF alternative that directly optimizes policies satisfying human preferences. DyMO~\citep{xie2025dymo} proposes a training-free alignment method for inference-time preference alignment. Flow-GRPO~\citep{liu2025flow} first integrates online reinforcement learning into flow matching models. However, Flow-GRPO applies uniform optimization across timesteps and suffers from sparse terminal rewards, ignoring time-varying exploration potential in stochastic processes. Our TempFlow-GRPO addresses these limitations through trajectory branching for precise credit assignment and noise-aware policy weighting aligned with natural exploration capacity.

\begin{figure}[t]
    \centering
    \begin{subfigure}[b]{0.48\textwidth}
        \begin{subfigure}[b]{\textwidth}
            \centering
            \includegraphics[width=\textwidth]{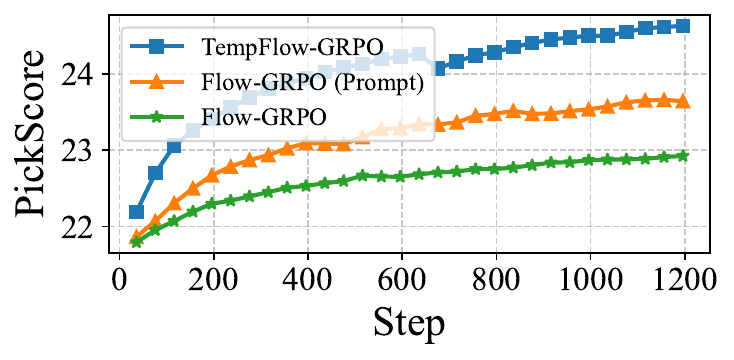}
            \label{fig2_1_1}
            \vspace{-0.4cm}
        \end{subfigure}
        \begin{subfigure}[b]{\textwidth}
            \centering
            \includegraphics[width=\textwidth]{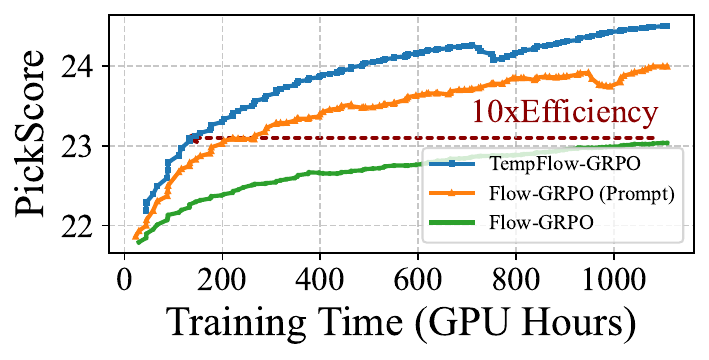}
            \label{fig2_1_2}
        \end{subfigure}
    \end{subfigure}
    \hfill
    \begin{subfigure}[b]{0.48\textwidth}
        \centering
        \includegraphics[width=\textwidth]{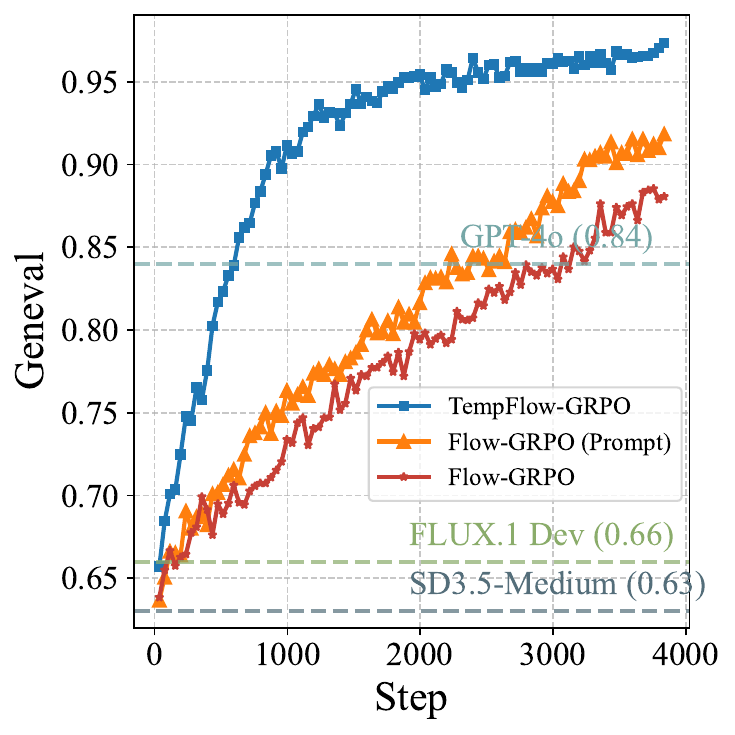}
        \label{fig2_2}
    \end{subfigure}
    \vspace{-0.5cm}
    \caption{(Left) Performance comparison on the PickScore benchmark across training steps and GPU hours. Flow-GRPO (Prompt) represents \textbf{an improved baseline with group-wise standard deviation stabilization}. TempFlow-GRPO consistently outperforms both Flow-GRPO variants in both sample efficiency (steps) and computational efficiency (GPU hours), \textbf{demonstrating superior training efficiency while achieving the best performance}. (Right) On the Geneval benchmark, TempFlow-GRPO achieves the highest performance, significantly outperforming Flow-GRPO and surpassing state-of-the-art models including GPT-4o, FLUX.1 Dev, and SD3.5-Medium.}
    \label{fig2}
    \vspace{-0.5cm}
\end{figure}

\noindent \textbf{Process Reward.} Shaping reward processes beyond sparse terminal rewards significantly improves learning and policy performance. Zhang et al.~\citep{zhang2025lessons} combines response-level and step-level evaluation metrics. ThinkPRM~\citep{khalifa2025process} builds data-efficient process reward models (PRMs) using verification chain-of-thought for step-wise verification. In diffusion models, SPO~\citep{liang2024step} trains step-aware preference models for both noisy and clean images. However, PRMs require expensive step-level supervision. Methods like PRIME~\citep{cui2025process} enable online PRM updates using only outcome labels through implicit process rewards. Math-Shepherd~\citep{wang2023math} assigns reward scores to solution steps. Given challenges in scoring noisy images, algorithms enabling process rewards for flow models are needed. Our TempFlow-GRPO circumvents specialized process reward models by directly attributing outcome-based signals to intermediate actions, enabling precise credit assignment without computational overhead of training step-level evaluators for semantically ambiguous states.

\section{Preliminary: Flow-GRPO}
\noindent \textbf{GRPO.} RL aims to learn a policy that maximizes the expected cumulative reward. GRPO optimizes the policy model by maximizing the following objective:
\begin{equation}
    \mathcal{J}_{\text{Flow-GRPO}}(\theta) = \mathbb{E}_{\bm{c}\sim\mathcal{C}, \{\bm{x}^i\}_{i=1}^G\sim\pi_{\theta_{\text{old}}}(\cdot|\bm{c})}f(r, \hat{A}, \theta, \epsilon, \beta)
\end{equation}
\begin{equation}
\begin{split}
    f(r, \hat{A}, \theta, \epsilon, \beta) & = \frac{1}{G} \sum \limits_{i=1}^{G} \frac{1}{T} \sum \limits_{t=0}^{T-1}(\mathop{min}(r_t^i(\theta)\hat{A}_t^i, \text{clip}(r_t^i(\theta), 1-\epsilon, 1+\epsilon)\hat{A}_t^i)-\beta D_{KL}(\pi_{\theta}||\pi_{\text{ref}})) \\
    r_t^i(\theta) &= \frac{p_{\theta}(\bm{x}^i_{t-1}|\bm{x}_t^i, \bm{c})}{p_{\theta_{\text{old}}}(\bm{x}^i_{t-1}|\bm{x}_t^i, \bm{c})}, T\text{ is the timestep.}
\end{split}
\end{equation}  
Given a prompt $\bm{c}$, the flow model $p_{\theta}$ samples a group of $G$ individual images $\{\bm{x}_0^i\}_{i=1}^G$ and the corresponding reverse-time trajectories $\{(\bm{x}_T^i, \bm{x}_{T-1}^i, ..., \bm{x}_0^i)\}_{i=1}^G$. Then, the advantage of the $i$-th image is calculated by normalizing the group-level rewards as follows:
\begin{equation}
    \hat{A}_t^i = \frac{R(\bm{x}_0^i, \bm{c}) - \text{mean}(\{R(\bm{x}^i_0,\bm{c})\}_{i=1}^G)}{\text{std}(\{R(\bm{x}_0^i,\bm{c})\}_{i=1}^G)}
\end{equation}
\noindent \textbf{Convert ODE to SDE.} Flow-GRPO converts the deterministic ODE into an equivalent SDE that matches the original model's marginal probability density function at all timesteps. The ODE and SDE is as follows:
\begin{equation}
    d\bm{x}_t = \bm{v}_tdt
\label{eq4}
\end{equation}
\begin{equation}
    \bm{x}_{t+\Delta t} = \bm{x}_t + [\bm{v}_{\theta}(\bm{x}_t, t) + \frac{\sigma_t^2}{2t}(\bm{x}_t+(1-t)\bm{v}_{\theta}(\bm{x}_t, t))]\Delta t + \sigma_t \sqrt{\Delta t} \bm{\epsilon}
\label{eq5}
\end{equation}
where $\bm{\epsilon} \sim \mathcal{N}(0,\bm{I})$ injects stochasticity and $\sigma_t = a\sqrt{\frac{t}{1-t}}$. And the KL divergence between $\pi_{\theta}$ and the reference policy $\pi_\text{ref}$ is a closed form:
\begin{equation}
    D_{KL}(\pi_{\theta}||\pi_{\text{ref}})=\frac{||\bar{\bm{x}}_{t+\Delta t,\theta}-\bar{\bm{x}}_{t+\Delta t,\text{ref}}||}{2\sigma_t^2\Delta t} = \frac{\Delta t}{2}(\frac{\sigma_t(1-t)}{2t}+\frac{1}{\sigma_t})^2||\bm{v}_{\theta}(\bm{x}_t, t) - \bm{v}_{\text{ref}}(\bm{x}_t, t)||^2
\end{equation}
\begin{figure*}[t]
    \centering
    \includegraphics[width=0.99\linewidth]{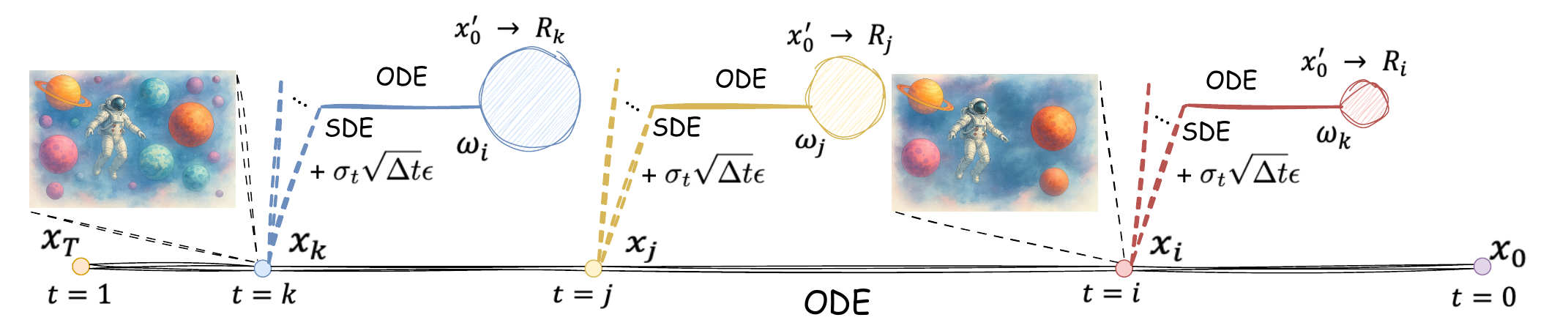}
    \caption{\textbf{Overview of TempFlow-GRPO Framework.} Our method performs trajectory branching by switching from ODE to SDE sampling at selected timesteps (t=k, j, i), injecting noise $\sigma_t\sqrt{\Delta t}\bm{\epsilon}$ to create exploratory branches. Each branch generates a distinct outcome with reward $R_i$, enabling precise credit assignment. The framework applies noise-aware weighting where $\omega_i > \omega_j > \omega_k$, prioritizing optimization at high-noise early stages (larger circles) over low-noise refinement phases (smaller circles), aligning learning intensity with each timestep's intrinsic exploration capacity. We visualize the model's learning process as an astronaut exploring unknown planets: in early stages , the model explores vast possibility spaces with high uncertainty, while later stages involve focused navigation toward the final destination.}
    \label{fig3}
    \vspace{-0.5cm}
\end{figure*} 
\section{Methods}
\subsection{Temporal Flow-GRPO}
Flow-GRPO advances online RL for flow matching but overlooks time-dependent generative dynamics. We identify two key limitations: \textbf{(1) Sparse Terminal Rewards}: uniform credit assignment across timesteps fails to distinguish critical early decisions from later fine-tuning; \textbf{(2) Uniform Optimization}: ignoring non-uniform noise in SDE sampling, where high-noise early steps offer greater exploration potential than low-noise refinement phases~\citep{xie2025dymo}. To address these issues, we propose \textbf{TempFlow-GRPO} (Figure~\ref{fig3}), introducing process rewards via trajectory branching and noise-aware policy weighting.

\subsubsection{Trajectory Branching for Process Rewards}
Traditional process reward methods require specialized reward models to evaluate noisy intermediate states $\bm{x}_t$, which is exceptionally difficult due to the semantic ambiguity of partially-denoised representations. We propose an elegant alternative that leverages the deterministic-stochastic sampling methods of flow matching models.

\textbf{Key Insight}: Instead of training complex process reward models, we use existing high-quality outcome-based reward models and directly attribute their scores to intermediate exploratory actions through a novel trajectory branching mechanism.

\textbf{Definition 1 (Trajectory Branching)}: We define a trajectory branching operation where a trajectory evolves deterministically until a \textbf{designated branching timestep $k$}, where $\bm{x}_k$ is obtained from initial noise $\bm{x}_T$ using Equation~\ref{eq4}. At this branching point, stochasticity is introduced via noise variable $\bm{\epsilon}$ in Equation~\ref{eq5}, yielding $\bm{x}_{k-1} = \text{SDE}(\bm{x}_k, \bm{\epsilon})$. The remainder of the trajectory $\bm{x}_{k-2}, \bm{x}_{k-3}, \ldots, \bm{x}_0$ is generated deterministically as $\bm{x}_0 = \text{ODE}^{k-1}(\bm{x}_{k-1})$, $\text{ODE}^{k-1}$ denotes $k-1$ times ODE.

\textbf{Theorem 1 (Credit Localization)}: Since all stochasticity and model controllability are concentrated at the branching point, the total reward variance and all parameter-dependent improvements are entirely attributable to the outcome of noise injection at $k$. This enables rigorous and efficient credit assignment localized to the branching point.\\
In practice, we replace the reward for the $k$-th step from $R(\bm{x}_0^i, \bm{c})$ to $R(\text{ODE}^{k-1}(\text{SDE}(\bm{x}_k^i, \bm{\epsilon}^i)), \bm{c})$, where $\bm{x}_0^i$ is sampled with SDE and $\text{ODE}^{k-1}(\text{SDE}(\bm{x}_k^i, \bm{\epsilon}^i))$ is sampled with ODE-SDE-ODE. Trajectory branching allows for the precise attribution of the terminal reward to step $k$, effectively creating a temporally-aware reward signal.

\begin{figure}[t]
    \centering
    \begin{subfigure}[b]{0.45\textwidth}
        \centering
        \includegraphics[width=\textwidth]{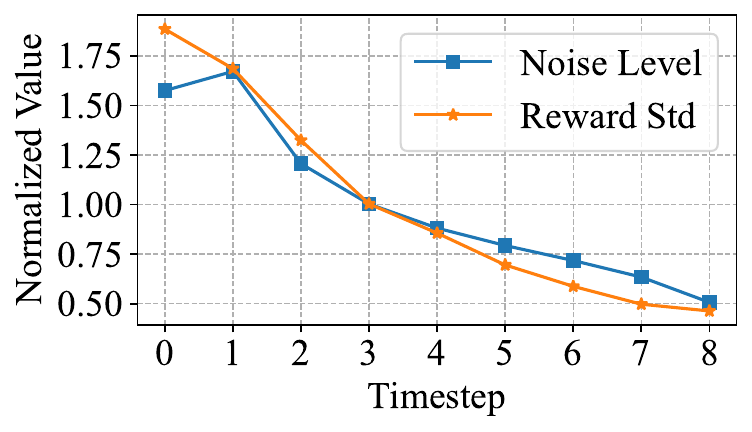}
        \label{fig4_1}
    \end{subfigure}
    \hfill
    \begin{subfigure}[b]{0.45\textwidth}
        \centering
        \includegraphics[width=\textwidth]{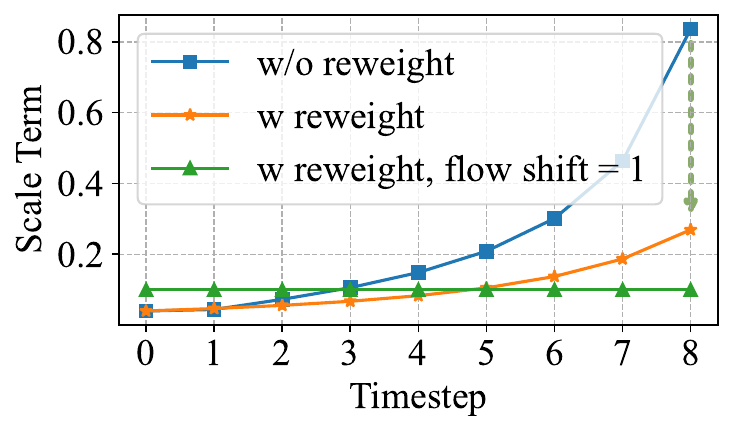}
        \label{fig4_2}
    \end{subfigure}
    \vspace{-0.8cm}
    \caption{\textbf{(Left)} Strong correlation between reward standard deviation and noise level across generative timesteps.  \textbf{(Right)} Scale term analysis reveals a fundamental mismatch in standard GRPO: scale terms are inversely proportional to noise levels, causing low-noise refinement steps to dominate optimization despite minimal impact on image content.    
    }
    \label{fig4}
    \vspace{-0.5cm}
\end{figure}

\subsubsection{Noise-Aware Policy Weighting}

While trajectory branching provides precise reward signals for individual timesteps, the generative process consists of a sequence of $T$ potential branching points with fundamentally different characteristics. The SDE sampler exhibits time-varying stochasticity: the noise injection magnitude $\sigma_t\sqrt{\Delta t}$ is large during initial generation stages and diminishes to near zero during final refinement stages. This non-uniform noise distribution implies that exploration capacity varies dramatically across timesteps. An exploratory action at an early stage has vastly different impact and risk compared to perturbations on near-perfect images. However, standard GRPO applies uniform optimization pressure, implicitly assuming equal learning importance at all stages.

\noindent \textbf{Reweighting by Noise Level.} We wonder if the level of the exploration space itself can serve as a proxy for a reweighting factor. We visualized the noise level alongside the reward standard deviation (Figure~\ref{fig4}, left) and observed a striking correspondence between the two. This strong correlation suggests that the noise level, serves as an excellent and intrinsic proxy for the exploration capacity and associated risk at each timestep.

We therefore propose to reweight the policy loss directly using the noise level. For each timestep $t$, we introduce a weighting factor proportional to the noise level. Specifically, we modify the original policy loss function to the following weighted form:
\begin{equation}
    \mathcal{J}_\text{policy}(\theta)=\frac{1}{G} \sum \limits_{i=1}^{G} \frac{1}{T} \sum \limits_{t=0}^{T-1}\text{Norm}(\sigma_t\sqrt{\Delta t})(\mathop{min}(r_t^i(\theta)\hat{A}_t^i, \text{clip}(r_t^i(\theta), 1-\epsilon, 1+\epsilon)\hat{A}_t^i)
\end{equation}
The intuition behind this weighting strategy is to align the optimization pressure with the inherent properties of the generative process. In the early stages of generation, noise is large, amplifying the learning signal during these high-noise, high-impact phases and encouraging the model to perform effective macroscopic exploration. As generation proceeds, noise diminishes, shifts the optimization focus towards fine-grained adjustments and stability, preventing aggressive exploration from corrupting a high-fidelity state with noise or artifacts.

\subsection{Policy Gradient-Based Theoretical Justification}
\label{sec4_2}
Consider a generative process parameterized by $\theta$, the policy gradient can be written as:
\begin{equation}
\nabla_\theta \mathcal{J}(\theta) = \sum_{k=0}^{T-1} \mathbb{E}_{\bm{x}_T \sim \mathcal{N}(0,\bm{I}), \bm{\epsilon} \sim \mathcal{N}(0,\bm{I})}[\nabla_\theta \log p_\theta(\bm{x}_{k-1}|\bm{x}_k) \hat{A}_k]
\label{eq8}
\end{equation}
This yields:
\begin{equation}
\nabla_\theta \mathcal{J}(\theta) = \sum_{k=0}^{T-1} \mathbb{E}_{\bm{x}_T \sim \mathcal{N}(0,\bm{I}), \bm{\epsilon} \sim \mathcal{N}(0,\bm{I})}\left[\left(\frac{1}{a} + \frac{a}{2}\right)\underbrace{\sqrt{\frac{\Delta k(1-k)}{k}}}_\text{Scale Term} \cdot \bm{\epsilon} \cdot \nabla_\theta \bm{v}_\theta(\bm{x}_k,k) \hat{A}_k\right]
\label{eq9}
\end{equation}
This reveals that the natural gradient coefficient is proportional to $\sqrt{\frac{\Delta k(1-k)}{k}}$, which captures the intrinsic exploration potential at timestep $k$. After reweighting, we have the following derivation:
\begin{equation}
\nabla_\theta \mathcal{J}(\theta) = \sum_{k=0}^{T-1} \mathbb{E}_{\bm{x}_T \sim \mathcal{N}(0,\bm{I}), \bm{\epsilon} \sim \mathcal{N}(0,\bm{I})}\left[\left(\frac{1}{a} + \frac{a}{2}\right)\underbrace{\Delta k}_\text{Scale Term} \cdot \bm{\epsilon} \cdot \nabla_\theta \bm{v}_\theta(\bm{x}_k,k) \hat{A}_k\right]
\label{eq10}
\end{equation}
where $\mathbb{E}_{\bm{\epsilon}}[\bm{\epsilon}\hat{A}_k] = \frac{g_k}{||g_k||}$, indicating that the norm of $\mathbb{E}_{\bm{\epsilon}}[\bm{\epsilon}\hat{A}_k]$ is invariant among the timesteps. Additional details about these equations are provided in Appendix~\ref{App1}. In Equation~\ref{eq9} and Equation~\ref{eq10}, the scale terms that modulate how each timestep's model gradient $\nabla_\theta \bm{v}_\theta(\bm{x}_k,k)$ contributes to the overall reward gradient simplify to distinct functions, which we denote as $\sqrt{\frac{\Delta k(1-k)}{k}}$ and $\Delta k$. We visualize these scale terms under different flow shifts in Figure~\ref{fig4} (right). The ``w/o reweighting" setting exhibits highly imbalanced gradient contributions across timesteps, where early denoising steps performing broad structural exploration contribute significantly less than late steps focused on fine-grained refinement. Our noise-aware policy reweighting mitigates this issue by simplifying the scale term to be proportional to step size $\Delta k$. When flow shift equals 1, our method achieves perfect equilibrium with equal gradient contributions from all timesteps, completely balancing the effect across the generation trajectory. \textbf{Additional details about this section are provided in Appendix~\ref{App1}}.

\subsection{Group Strategy}
\label{sec_4_3}
\noindent \textbf{Batch Group \& Prompt Group.} The original GRPO~\citep{shao2024deepseekmath} framework employs a group-level strategy, where trajectories are grouped based on a shared prompt. To mitigate issues of overfitting and instability, Reinforce++~\citep{hu2025reinforce++} later introduced a batch-level normalization.

\begin{wrapfigure}{r}{0.35\textwidth}
    \centering
    \includegraphics[width=0.33\textwidth]{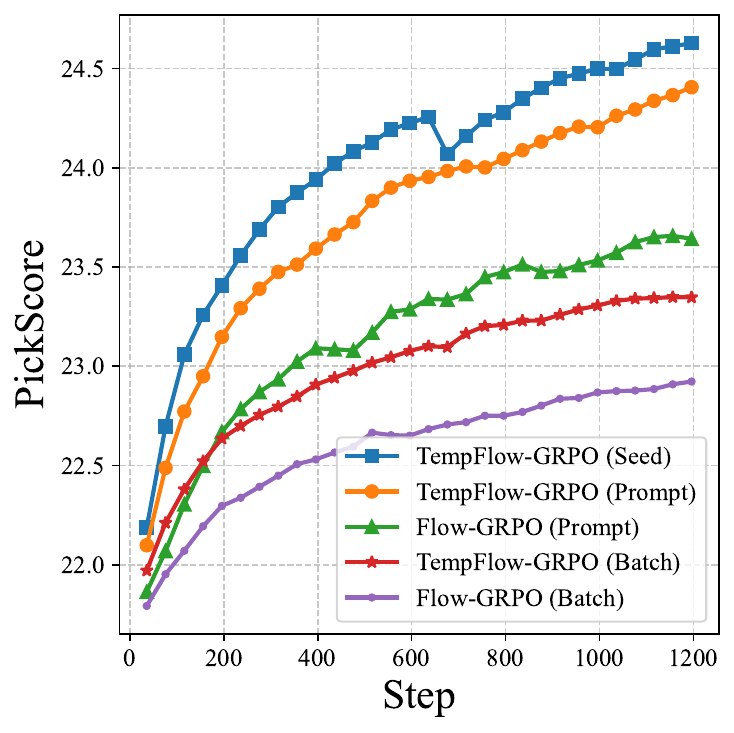}
    \vspace{-0.2cm}
    \caption{Performance of TempFlow-GRPO and Flow-GRPO on different group strategy.}
\label{fig5}
\vspace{-0.5cm}
\end{wrapfigure}

\noindent \textbf{Seed Group.} In our work, trajectory branching requires $K$ distinct explorations at each timestep. Furthermore, we propose a novel seed-level grouping strategy. Under this approach, trajectories originating from the same prompt are further grouped if they share an identical initial noise. This methodology effectively controls for the influence of the initial noise, thereby ensuring that variations in reward can be attributed solely to the exploration conducted during the branching process. The experimental results, presented in Figure~\ref{fig5}, validate our approach. As shown, TempFlow-GRPO consistently achieves superior performance compared to Flow-GRPO, regardless of the grouping strategy employed. 

\section{Experiment}
Following Flow-GRPO, we validate our approach on the Compositional Image Generation benchmark in Geneval~\citep{ghosh2023geneval} and the Human Preference Alignment task in PickScore~\citep{kirstain2023pick}. To ensure a fair comparison, our experimental setup is kept consistent with that of Flow-GRPO. For instance, we normalized the weights applied to the policy loss to have a mean of 1 at all timesteps. While our method introduces a trajectory branching strategy, we maintain an identical group configuration by setting 4 initial noise seeds and a branching factor ($K$) of 6. This results in a total group size of 24 (4 seeds $\times$ 6 branches) and 48 groups, perfectly matching the setup in Flow-GRPO. Furthermore, to benchmark our method against concurrent work, we conduct additional experiments on the FLUX.1-dev model. \textbf{More detailed experimental settings (include comparsion with DanceGRPO) can be found in the Appendix~\ref{app2}.}

\subsection{Main Results}
\begin{table*}[t]
    \centering
    \caption{{\bf GenEval Result.} 
Best scores are in \colorbox{pearDark!20}{blue}, second-best in \colorbox{color_green}{green}. Results for models are from Flow-GRPO. Obj.: Object; Attr.: Attribution.}
    \resizebox{\linewidth}{!}{
    \begin{tabular}{l|c|c|cccccc}
    \toprule
    \textbf{Model} & \textbf{Step}  & \textbf{Overall $\uparrow$ } & \textbf{Single Obj. $\uparrow$} & \textbf{Two Obj. $\uparrow$} & \textbf{Counting $\uparrow$ } & \textbf{Colors $\uparrow$} & \textbf{Position $\uparrow$} & \textbf{Attr. Binding $\uparrow$} \\
    \midrule
    \multicolumn{9}{c}{\textit{Diffusion Models}} \\
    \midrule
    LDM~\citep{rombach2022high}  & - & 0.37 & 0.92 & 0.29 & 0.23 & 0.70 & 0.02 & 0.05 \\
    SD1.5~\citep{rombach2022high}  & - & 0.43 & 0.97 & 0.38 & 0.35 & 0.76 & 0.04 & 0.06 \\
    SD2.1~\citep{rombach2022high}  & - & 0.50 & 0.98 & 0.51 & 0.44 & 0.85 & 0.07 & 0.17 \\
    SD-XL~\citep{podell2023sdxl}  & - & 0.55 & 0.98 & 0.74 & 0.39 & 0.85 & 0.15 & 0.23 \\
    DALLE-2~\citep{ramesh2022hierarchical}  & - & 0.52 & 0.94 & 0.66 & 0.49 & 0.77 & 0.10 & 0.19 \\
    DALLE-3~\citep{betker2023improving}  & - & 0.67 & 0.96 & 0.87 & 0.47 & 0.83 & 0.43 & 0.45 \\
    \midrule
    \multicolumn{9}{c}{\textit{Autoregressive Models}} \\
    \midrule
    Show-o~\citep{xie2024show}  & - & 0.53 & 0.95 & 0.52 & 0.49 & 0.82 & 0.11 & 0.28 \\
    Emu3-Gen~\citep{wang2024emu3}  & - & 0.54 & 0.98 & 0.71 & 0.34 & 0.81 & 0.17 & 0.21 \\
    JanusFlow~\citep{ma2025janusflow}  & - & 0.63 & 0.97 & 0.59 & 0.45 & 0.83 & 0.53 & 0.42 \\
    Janus-Pro-7B~\citep{chen2025janus}  & - & 0.80 & \colorbox{color_green}{0.99} & 0.89 & 0.59 & 0.90 & 0.79 & 0.66 \\
    GPT-4o~\citep{hurst2024gpt} & - & 0.84 & \colorbox{color_green}{0.99} & 0.92 & 0.85 & \colorbox{color_green}{0.92} & 0.75 & 0.61 \\
    \midrule
    \multicolumn{9}{c}{\textit{Flow Matching Models}} \\
    \midrule
    FLUX.1 Dev~\citep{black2025flux}  & - & 0.66 & 0.98 & 0.81 & 0.74 & 0.79 & 0.22 & 0.45 \\
    SD3.5-L~\citep{esser2024scaling}  & - & 0.71 & 0.98 & 0.89 & 0.73 & 0.83 & 0.34 & 0.47 \\
    SANA-1.5 4.8B~\citep{xie2025sana} & - & 0.81 & \colorbox{color_green}{0.99} & 0.93 & 0.86 & 0.84 & 0.59 & 0.65 \\
    SD3.5-M~\citep{esser2024scaling} & - & 0.63 & 0.98 & 0.78 & 0.50 & 0.81 & 0.24 & 0.52 \\
    \midrule
    \multicolumn{9}{c}{\textit{GRPO based Methods}} \\
    \midrule
    SD3.5-M+Flow-GRPO~\citep{liu2025flow} & 5600 & \colorbox{color_green}{0.95} & \colorbox{pearDark!20}{1.00} & \colorbox{color_green}{0.99} & \colorbox{color_green}{0.95} & \colorbox{color_green}{0.92} & \colorbox{pearDark!20}{0.99} & \colorbox{color_green}{0.86}  \\
    SD3.5-M+Flow-GRPO~\citep{liu2025flow} & 3800 & 0.88 & \colorbox{color_green}{0.99} & 0.96 & 0.90 & 0.87 & \colorbox{color_green}{0.83} & 0.78  \\
    \textbf{SD3.5-M+TempFlow-GRPO} & 3800 & \colorbox{pearDark!20}{0.97} & \colorbox{pearDark!20}{1.00} & \colorbox{pearDark!20}{1.00} & \colorbox{pearDark!20}{0.96} & \colorbox{pearDark!20}{0.95} & \colorbox{pearDark!20}{0.99} & \colorbox{pearDark!20}{0.91}  \\
    \bottomrule
    \end{tabular}
    }
    \label{tab1}
\vspace{-0.5cm}
\end{table*}

\noindent \textbf{Compositional Image Generation. }We evaluate the compositional image generation capability of TempFlow-GRPO on the Geneval benchmark with its corresponding reward model. The experimental results are summarized in Table~\ref{tab1}. As shown, our approach significantly improves the performance of the base model, increasing the overall score from 0.63 to 0.97. Furthermore, among GRPO-based methods, our method substantially outperforms Flow-GRPO: it achieves a performance of 0.97 within only 3,800 steps, whereas Flow-GRPO reaches only 0.88 under the same conditions. Additionally, as illustrated in Figure~\ref{fig2}, our method requires only about 2,000 steps to achieve a score of 0.95, while Flow-GRPO needs approximately 5,600 steps to reach the same level. Overall, these results demonstrate that TempFlow-GRPO not only accelerates convergence but also achieves superior final performance compared to existing approaches.

\noindent \textbf{Human Preference Alignment. }To further validate the generalizability of our approach, we conducted experiments on the PickScore benchmark, using PickScore as the reward model. As shown in Figure~\ref{fig2} (left), TempFlow-GRPO, achieves the highest performance, surpassing the original Flow-GRPO by approximately 1.7\% and outperforming the improved baseline Flow-GRPO (Prompt) by about 1.0\%. Notably, our method requires only 100–200 training steps to match the performance of Flow-GRPO, and just 300–400 steps to reach the level of Flow-GRPO (Prompt). These results on PickScore further demonstrate the general applicability of our method as a unified flow-based RL algorithm across different reward models.

\begin{figure}[htbp]
    \centering
    \begin{subfigure}[b]{0.48\textwidth}
        \centering
        \includegraphics[width=\textwidth]{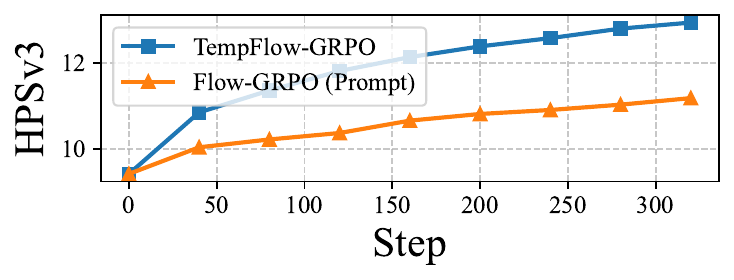}
        \label{fig6_1}
    \end{subfigure}
    \hfill
    \begin{subfigure}[b]{0.48\textwidth}
        \centering
        \includegraphics[width=\textwidth]{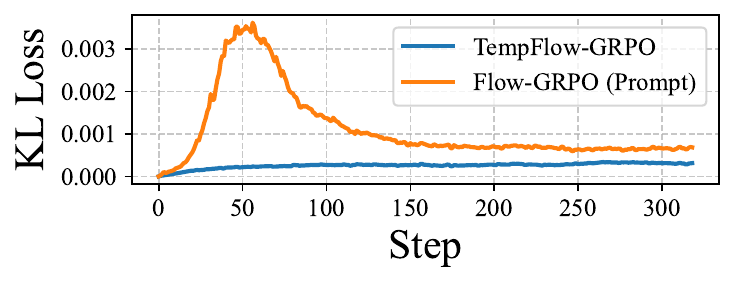}
        \label{fig6_2}
    \end{subfigure}
    \vspace{-0.8cm}
    \caption{The performance and KL loss (smoothed) of HPSv3 on PickScore dataset (FLUX.1-dev).}
    \label{fig6}
    \vspace{-0.3cm}
\end{figure}
\noindent \textbf{HPSv3 on FLUX.1-dev.} To further validate the performance of TempFlow-GRPO across different model and reward models, we select FLUX.1-dev as our base model with 1024 resolution and employ HPSv3~\citep{ma2025hpsv3} as the reward model.  As shown in Figure~\ref{fig6}, which presents the results, our method requires only 80 steps to match the performance that Flow-GPO achieves in 300 steps. Moreover, our approach maintains a lower and more stable KL loss. \textbf{Further experimental results and visualizations for the FLUX model can be found in the Appendix.~\ref{app3}.}

\subsection{Analysis}

\begin{figure}[t]
    \centering
    \begin{subfigure}[b]{0.32\textwidth}
        \centering
        \includegraphics[width=\textwidth]{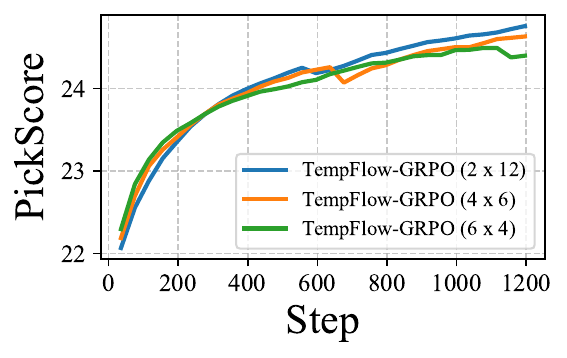}
        \label{fig7_1}
    \end{subfigure}
    \begin{subfigure}[b]{0.32\textwidth}
        \centering
        \includegraphics[width=\textwidth]{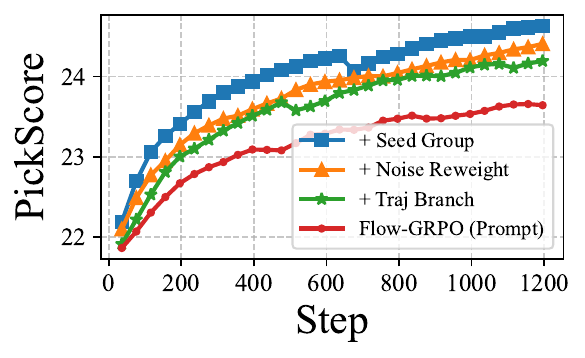}
        \label{fig7_2}
    \end{subfigure}
    \begin{subfigure}[b]{0.32\textwidth}
        \centering
        \includegraphics[width=\textwidth]{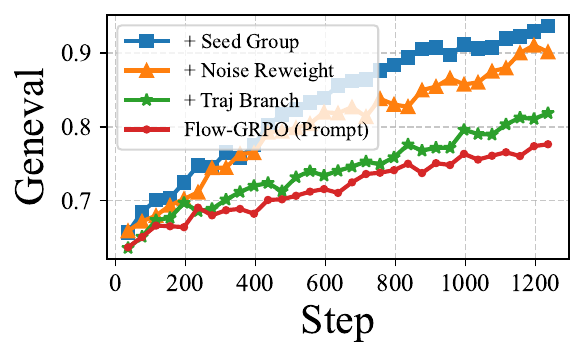}
        \label{fig7_3}
    \end{subfigure}
    \vspace{-0.8cm}
    \caption{Ablation studies on trajectory branch and noise-aware policy reweighting.}
    \label{fig7}
    \vspace{-0.5cm}
\end{figure}

\noindent \textbf{Ablation of Trajectory branching. }In our trajectory branching strategy, branching is performed at each step. To ensure a fair comparison with Flow-GRPO, we maintain a constant total group size of 24. Therefore, we tested several branching configurations: 2 initial noises with 12 branches each (2$\times$12), 4 initial noises with 6 branches each (4$\times$6), and 6 initial noises with 4 branches each (6$\times$4). The experimental results are shown in the left of Figure~\ref{fig7}. The figure indicates that a larger number of initial noises helps to accelerate performance improvement in the early stages of training. However, as training progresses, a higher number of branches yields better results. To strike a balance, we ultimately selected the 4$\times$6 configuration as the default setting for our paper.

\noindent \textbf{Ablation of TempFlow-GRPO. }We conducted ablation studies to explore the effectiveness of our proposed components: the trajectory branch, noise-aware policy weighting and seed group. These ablations were performed on both the Geneval and PickScore benchmarks. As shown in the second and third subfigures of  Figure~\ref{fig7}, on the PickScore benchmark, introducing the trajectory branch further improves the performance of Flow-GRPO (Prompt), and applying noise-aware reweighting on top of this further boosted the performance. Finally, we achieved the highest performance by applying seed group. On the Geneval benchmark, the benefit of the noise-aware strategy is even more significant: compared to Flow-GRPO, noise-aware policy reweighting boosts performance from 0.82 to 0.92 in 1200 step,  a 10\% improvement while the trajectory branch also brings about a substantial gain of approximately 5\% and seed group achieves 2\% improvement. These ablation results clearly demonstrate the effectiveness of our proposed methods. 

\begin{figure*}[t]
    \centering
    \includegraphics[width=\linewidth]{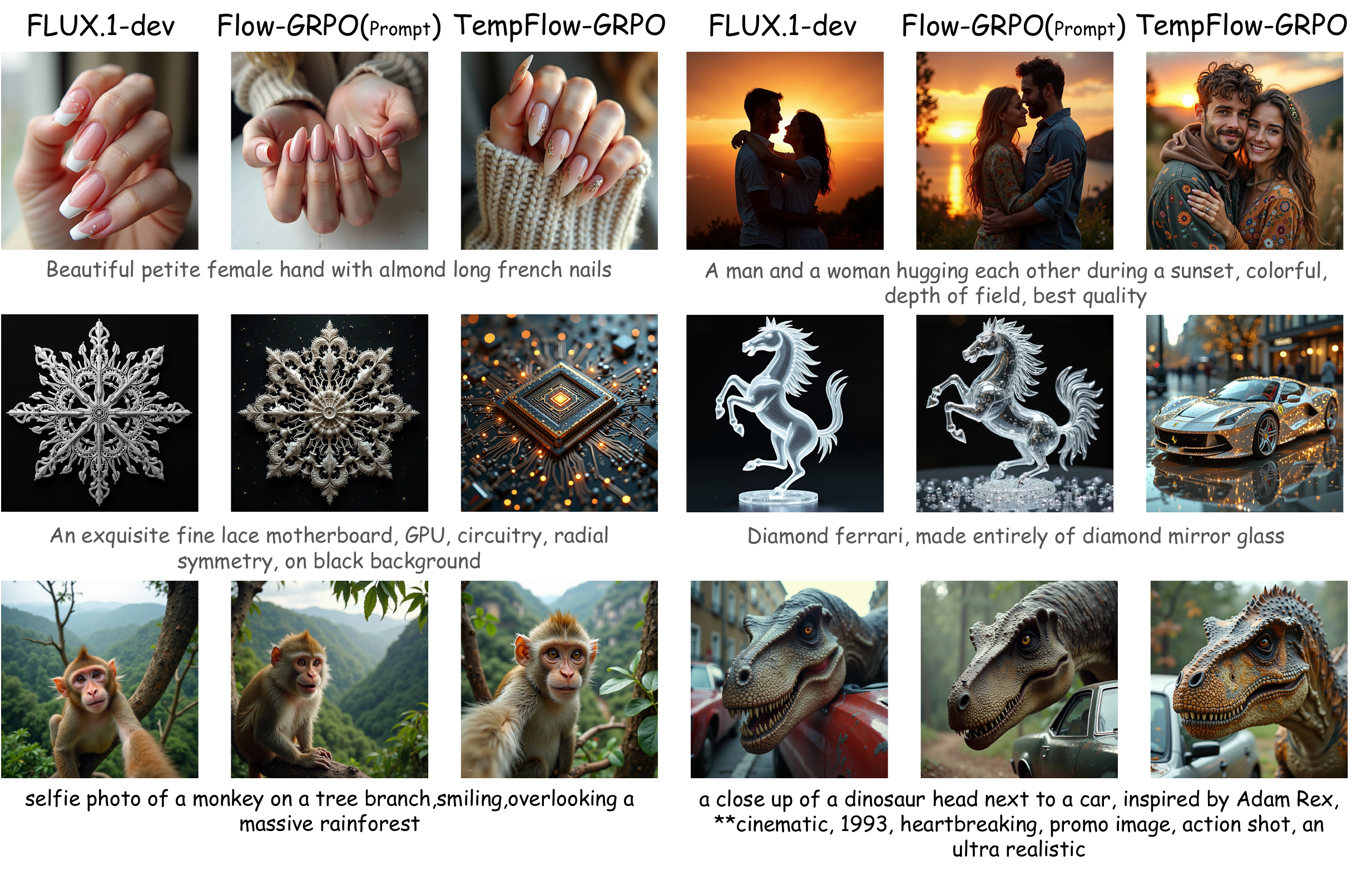}
    \vspace{-0.8cm}
    \caption{Qualitative comparison between FLUX.1-dev, Flow-GRPO (Prompt) and TempFlow-GRPO with HPSv3 as reward on PickScore prompts.}
    \label{fig8}
    \vspace{-0.8cm}
\end{figure*}

\noindent \textbf{Qualitative Result. }We also conducted qualitative analyses on the FLUX.1-dev, Flow-GRPO (Prompt), and TempFlow-GRPO. As shown in Figure~\ref{fig8}, compared to Flow-GRPO (Prompt), TempFlow-GRPO produces images with noticeably finer details and fewer visual artifacts or mistakes. In particular, our approach demonstrates superior capability in preserving complex structures and realistic textures. These qualitative improvements further highlight the advantages of our method in generating high-quality, visually appealing images.
\vspace{-0.5cm}

\section{Conclusion}
\vspace{-0.3cm}
We presented TempFlow-GRPO, a temporally-aware reinforcement learning framework that addresses fundamental limitations in existing flow-based GRPO methods. Through trajectory branching, we enable precise credit assignment to intermediate actions without requiring specialized process reward models. Through noise-aware weighting, we ensure that optimization intensity matches each timestep's exploration potential. We also introduce a principled seed group strategy that controls for initialization effects. Our extensive experiments demonstrate that TempFlow-GRPO achieves SOTA performance on human preference.

\noindent \textbf{Limitations.} Although our method achieves significant improvements in both performance and image quality, the current experiments focus primarily on algorithmic innovations rather than reward model enhancements. \textbf{In future work, we plan to explore how to effectively incorporate multi-modal rewards from more powerful foundation models and develop comprehensive reward frameworks, aiming to enhance performance across multiple evaluation dimensions.}

\noindent\textbf{Use of LLMs. }We utilize LLMs to assist with formula derivations and writing refinement.


\bibliography{arxiv}

\begin{thebibliography}{36}
\providecommand{\natexlab}[1]{#1}
\providecommand{\url}[1]{\texttt{#1}}
\expandafter\ifx\csname urlstyle\endcsname\relax
  \providecommand{\doi}[1]{doi: #1}\else
  \providecommand{\doi}{doi: \begingroup \urlstyle{rm}\Url}\fi

\bibitem[Betker et~al.(2023)Betker, Goh, Jing, Brooks, Wang, Li, Ouyang, Zhuang, Lee, Guo, et~al.]{betker2023improving}
James Betker, Gabriel Goh, Li~Jing, Tim Brooks, Jianfeng Wang, Linjie Li, Long Ouyang, Juntang Zhuang, Joyce Lee, Yufei Guo, et~al.
\newblock Improving image generation with better captions.
\newblock \emph{Computer Science. https://cdn. openai. com/papers/dall-e-3. pdf}, 2\penalty0 (3):\penalty0 8, 2023.

\bibitem[Black et~al.(2025)]{black2025flux}
Forest~Labs Black et~al.
\newblock Flux.1 kontext: Flow matching for in-context image generation and editing in latent space.
\newblock \emph{arXiv preprint arXiv:2506.15742}, 2025.

\bibitem[Black et~al.(2023)Black, Janner, Du, Kostrikov, and Levine]{black2023training}
Kevin Black, Michael Janner, Yilun Du, Ilya Kostrikov, and Sergey Levine.
\newblock Training diffusion models with reinforcement learning.
\newblock \emph{arXiv preprint arXiv:2305.13301}, 2023.

\bibitem[Chen et~al.(2025)Chen, Wu, Liu, Pan, Liu, Xie, Yu, and Ruan]{chen2025janus}
Xiaokang Chen, Zhiyu Wu, Xingchao Liu, Zizheng Pan, Wen Liu, Zhenda Xie, Xingkai Yu, and Chong Ruan.
\newblock Janus-pro: Unified multimodal understanding and generation with data and model scaling.
\newblock \emph{arXiv preprint arXiv:2501.17811}, 2025.

\bibitem[Cui et~al.(2025)Cui, Yuan, Wang, Wang, Li, He, Fan, Yu, Xu, Chen, et~al.]{cui2025process}
Ganqu Cui, Lifan Yuan, Zefan Wang, Hanbin Wang, Wendi Li, Bingxiang He, Yuchen Fan, Tianyu Yu, Qixin Xu, Weize Chen, et~al.
\newblock Process reinforcement through implicit rewards.
\newblock \emph{arXiv preprint arXiv:2502.01456}, 2025.

\bibitem[Esser et~al.(2024)Esser, Kulal, Blattmann, Entezari, M{\"u}ller, Saini, Levi, Lorenz, Sauer, Boesel, et~al.]{esser2024scaling}
Patrick Esser, Sumith Kulal, Andreas Blattmann, Rahim Entezari, Jonas M{\"u}ller, Harry Saini, Yam Levi, Dominik Lorenz, Axel Sauer, Frederic Boesel, et~al.
\newblock Scaling rectified flow transformers for high-resolution image synthesis.
\newblock In \emph{Forty-first international conference on machine learning}, 2024.

\bibitem[Fan et~al.(2023)Fan, Watkins, Du, Liu, Ryu, Boutilier, Abbeel, Ghavamzadeh, Lee, and Lee]{fan2023dpok}
Ying Fan, Olivia Watkins, Yuqing Du, Hao Liu, Moonkyung Ryu, Craig Boutilier, Pieter Abbeel, Mohammad Ghavamzadeh, Kangwook Lee, and Kimin Lee.
\newblock Dpok: Reinforcement learning for fine-tuning text-to-image diffusion models.
\newblock \emph{Advances in Neural Information Processing Systems}, 36:\penalty0 79858--79885, 2023.

\bibitem[Ghosh et~al.(2023)Ghosh, Hajishirzi, and Schmidt]{ghosh2023geneval}
Dhruba Ghosh, Hannaneh Hajishirzi, and Ludwig Schmidt.
\newblock Geneval: An object-focused framework for evaluating text-to-image alignment.
\newblock \emph{Advances in Neural Information Processing Systems}, 36:\penalty0 52132--52152, 2023.

\bibitem[Gu et~al.(2024)Gu, Wang, Yin, Xie, and Zhou]{gu2024diffusion}
Yi~Gu, Zhendong Wang, Yueqin Yin, Yujia Xie, and Mingyuan Zhou.
\newblock Diffusion-rpo: Aligning diffusion models through relative preference optimization.
\newblock \emph{arXiv preprint arXiv:2406.06382}, 2024.

\bibitem[Hong et~al.(2024)Hong, Paul, Lee, Rasul, Thorne, and Jeong]{hong2024margin}
Jiwoo Hong, Sayak Paul, Noah Lee, Kashif Rasul, James Thorne, and Jongheon Jeong.
\newblock Margin-aware preference optimization for aligning diffusion models without reference.
\newblock In \emph{First Workshop on Scalable Optimization for Efficient and Adaptive Foundation Models}, 2024.

\bibitem[Hu et~al.(2025)Hu, Liu, Xu, and Shen]{hu2025reinforce++}
Jian Hu, Jason~Klein Liu, Haotian Xu, and Wei Shen.
\newblock Reinforce++: An efficient rlhf algorithm with robustness to both prompt and reward models.
\newblock \emph{arXiv preprint arXiv:2501.03262}, 2025.

\bibitem[Hurst et~al.(2024)Hurst, Lerer, Goucher, Perelman, Ramesh, Clark, Ostrow, Welihinda, Hayes, Radford, et~al.]{hurst2024gpt}
Aaron Hurst, Adam Lerer, Adam~P Goucher, Adam Perelman, Aditya Ramesh, Aidan Clark, AJ~Ostrow, Akila Welihinda, Alan Hayes, Alec Radford, et~al.
\newblock Gpt-4o system card.
\newblock \emph{arXiv preprint arXiv:2410.21276}, 2024.

\bibitem[Khalifa et~al.(2025)Khalifa, Agarwal, Logeswaran, Kim, Peng, Lee, Lee, and Wang]{khalifa2025process}
Muhammad Khalifa, Rishabh Agarwal, Lajanugen Logeswaran, Jaekyeom Kim, Hao Peng, Moontae Lee, Honglak Lee, and Lu~Wang.
\newblock Process reward models that think.
\newblock \emph{arXiv preprint arXiv:2504.16828}, 2025.

\bibitem[Kirstain et~al.(2023)Kirstain, Polyak, Singer, Matiana, Penna, and Levy]{kirstain2023pick}
Yuval Kirstain, Adam Polyak, Uriel Singer, Shahbuland Matiana, Joe Penna, and Omer Levy.
\newblock Pick-a-pic: An open dataset of user preferences for text-to-image generation.
\newblock \emph{Advances in neural information processing systems}, 36:\penalty0 36652--36663, 2023.

\bibitem[Labs et~al.(2025)Labs, Batifol, Blattmann, Boesel, Consul, Diagne, Dockhorn, English, English, Esser, et~al.]{labs2025flux}
Black~Forest Labs, Stephen Batifol, Andreas Blattmann, Frederic Boesel, Saksham Consul, Cyril Diagne, Tim Dockhorn, Jack English, Zion English, Patrick Esser, et~al.
\newblock Flux. 1 kontext: Flow matching for in-context image generation and editing in latent space.
\newblock \emph{arXiv preprint arXiv:2506.15742}, 2025.

\bibitem[Liang et~al.(2024)Liang, Yuan, Gu, Chen, Hang, Li, and Zheng]{liang2024step}
Zhanhao Liang, Yuhui Yuan, Shuyang Gu, Bohan Chen, Tiankai Hang, Ji~Li, and Liang Zheng.
\newblock Step-aware preference optimization: Aligning preference with denoising performance at each step.
\newblock \emph{arXiv preprint arXiv:2406.04314}, 2\penalty0 (5):\penalty0 7, 2024.

\bibitem[Liang et~al.(2025)Liang, Yuan, Gu, Chen, Hang, Cheng, Li, and Zheng]{liang2025aesthetic}
Zhanhao Liang, Yuhui Yuan, Shuyang Gu, Bohan Chen, Tiankai Hang, Mingxi Cheng, Ji~Li, and Liang Zheng.
\newblock Aesthetic post-training diffusion models from generic preferences with step-by-step preference optimization.
\newblock In \emph{Proceedings of the Computer Vision and Pattern Recognition Conference}, pp.\  13199--13208, 2025.

\bibitem[Liu et~al.(2025)Liu, Liu, Liang, Li, Liu, Wang, Wan, Zhang, and Ouyang]{liu2025flow}
Jie Liu, Gongye Liu, Jiajun Liang, Yangguang Li, Jiaheng Liu, Xintao Wang, Pengfei Wan, Di~Zhang, and Wanli Ouyang.
\newblock Flow-grpo: Training flow matching models via online rl.
\newblock \emph{arXiv preprint arXiv:2505.05470}, 2025.

\bibitem[Ma et~al.(2025{\natexlab{a}})Ma, Liu, Chen, Liu, Wu, Wu, Pan, Xie, Zhang, Yu, et~al.]{ma2025janusflow}
Yiyang Ma, Xingchao Liu, Xiaokang Chen, Wen Liu, Chengyue Wu, Zhiyu Wu, Zizheng Pan, Zhenda Xie, Haowei Zhang, Xingkai Yu, et~al.
\newblock Janusflow: Harmonizing autoregression and rectified flow for unified multimodal understanding and generation.
\newblock In \emph{Proceedings of the Computer Vision and Pattern Recognition Conference}, pp.\  7739--7751, 2025{\natexlab{a}}.

\bibitem[Ma et~al.(2025{\natexlab{b}})Ma, Wu, Sun, and Li]{ma2025hpsv3}
Yuhang Ma, Xiaoshi Wu, Keqiang Sun, and Hongsheng Li.
\newblock Hpsv3: Towards wide-spectrum human preference score.
\newblock \emph{arXiv preprint arXiv:2508.03789}, 2025{\natexlab{b}}.

\bibitem[Podell et~al.(2023)Podell, English, Lacey, Blattmann, Dockhorn, M{\"u}ller, Penna, and Rombach]{podell2023sdxl}
Dustin Podell, Zion English, Kyle Lacey, Andreas Blattmann, Tim Dockhorn, Jonas M{\"u}ller, Joe Penna, and Robin Rombach.
\newblock Sdxl: Improving latent diffusion models for high-resolution image synthesis.
\newblock \emph{arXiv preprint arXiv:2307.01952}, 2023.

\bibitem[Rafailov et~al.(2023)Rafailov, Sharma, Mitchell, Manning, Ermon, and Finn]{rafailov2023direct}
Rafael Rafailov, Archit Sharma, Eric Mitchell, Christopher~D Manning, Stefano Ermon, and Chelsea Finn.
\newblock Direct preference optimization: Your language model is secretly a reward model.
\newblock \emph{Advances in neural information processing systems}, 36:\penalty0 53728--53741, 2023.

\bibitem[Ramesh et~al.(2022)Ramesh, Dhariwal, Nichol, Chu, and Chen]{ramesh2022hierarchical}
Aditya Ramesh, Prafulla Dhariwal, Alex Nichol, Casey Chu, and Mark Chen.
\newblock Hierarchical text-conditional image generation with clip latents.
\newblock \emph{arXiv preprint arXiv:2204.06125}, 1\penalty0 (2):\penalty0 3, 2022.

\bibitem[Rombach et~al.(2022)Rombach, Blattmann, Lorenz, Esser, and Ommer]{rombach2022high}
Robin Rombach, Andreas Blattmann, Dominik Lorenz, Patrick Esser, and Bj{\"o}rn Ommer.
\newblock High-resolution image synthesis with latent diffusion models.
\newblock In \emph{Proceedings of the IEEE/CVF conference on computer vision and pattern recognition}, pp.\  10684--10695, 2022.

\bibitem[Shao et~al.(2024)Shao, Wang, Zhu, Xu, Song, Bi, Zhang, Zhang, Li, Wu, et~al.]{shao2024deepseekmath}
Zhihong Shao, Peiyi Wang, Qihao Zhu, Runxin Xu, Junxiao Song, Xiao Bi, Haowei Zhang, Mingchuan Zhang, YK~Li, Yang Wu, et~al.
\newblock Deepseekmath: Pushing the limits of mathematical reasoning in open language models.
\newblock \emph{arXiv preprint arXiv:2402.03300}, 2024.

\bibitem[Wallace et~al.(2024)Wallace, Dang, Rafailov, Zhou, Lou, Purushwalkam, Ermon, Xiong, Joty, and Naik]{wallace2024diffusion}
Bram Wallace, Meihua Dang, Rafael Rafailov, Linqi Zhou, Aaron Lou, Senthil Purushwalkam, Stefano Ermon, Caiming Xiong, Shafiq Joty, and Nikhil Naik.
\newblock Diffusion model alignment using direct preference optimization.
\newblock In \emph{Proceedings of the IEEE/CVF Conference on Computer Vision and Pattern Recognition}, pp.\  8228--8238, 2024.

\bibitem[Wang et~al.(2023)Wang, Li, Shao, Xu, Dai, Li, Chen, Wu, and Sui]{wang2023math}
Peiyi Wang, Lei Li, Zhihong Shao, RX~Xu, Damai Dai, Yifei Li, Deli Chen, Yu~Wu, and Zhifang Sui.
\newblock Math-shepherd: Verify and reinforce llms step-by-step without human annotations.
\newblock \emph{arXiv preprint arXiv:2312.08935}, 2023.

\bibitem[Wang et~al.(2024)Wang, Zhang, Luo, Sun, Cui, Wang, Zhang, Wang, Li, Yu, et~al.]{wang2024emu3}
Xinlong Wang, Xiaosong Zhang, Zhengxiong Luo, Quan Sun, Yufeng Cui, Jinsheng Wang, Fan Zhang, Yueze Wang, Zhen Li, Qiying Yu, et~al.
\newblock Emu3: Next-token prediction is all you need.
\newblock \emph{arXiv preprint arXiv:2409.18869}, 2024.

\bibitem[Wu et~al.(2023)Wu, Hao, Sun, Chen, Zhu, Zhao, and Li]{wu2023human}
Xiaoshi Wu, Yiming Hao, Keqiang Sun, Yixiong Chen, Feng Zhu, Rui Zhao, and Hongsheng Li.
\newblock Human preference score v2: A solid benchmark for evaluating human preferences of text-to-image synthesis.
\newblock \emph{arXiv preprint arXiv:2306.09341}, 2023.

\bibitem[Xie et~al.(2024{\natexlab{a}})Xie, Chen, Chen, Cai, Tang, Lin, Zhang, Li, Zhu, Lu, et~al.]{xie2024sana}
Enze Xie, Junsong Chen, Junyu Chen, Han Cai, Haotian Tang, Yujun Lin, Zhekai Zhang, Muyang Li, Ligeng Zhu, Yao Lu, et~al.
\newblock Sana: Efficient high-resolution image synthesis with linear diffusion transformers.
\newblock \emph{arXiv preprint arXiv:2410.10629}, 2024{\natexlab{a}}.

\bibitem[Xie et~al.(2025)Xie, Chen, Zhao, Yu, Zhu, Wu, Lin, Zhang, Li, Chen, et~al.]{xie2025sana}
Enze Xie, Junsong Chen, Yuyang Zhao, Jincheng Yu, Ligeng Zhu, Chengyue Wu, Yujun Lin, Zhekai Zhang, Muyang Li, Junyu Chen, et~al.
\newblock Sana 1.5: Efficient scaling of training-time and inference-time compute in linear diffusion transformer.
\newblock \emph{arXiv preprint arXiv:2501.18427}, 2025.

\bibitem[Xie et~al.(2024{\natexlab{b}})Xie, Mao, Bai, Zhang, Wang, Lin, Gu, Chen, Yang, and Shou]{xie2024show}
Jinheng Xie, Weijia Mao, Zechen Bai, David~Junhao Zhang, Weihao Wang, Kevin~Qinghong Lin, Yuchao Gu, Zhijie Chen, Zhenheng Yang, and Mike~Zheng Shou.
\newblock Show-o: One single transformer to unify multimodal understanding and generation.
\newblock \emph{arXiv preprint arXiv:2408.12528}, 2024{\natexlab{b}}.

\bibitem[Xie \& Gong(2025)Xie and Gong]{xie2025dymo}
Xin Xie and Dong Gong.
\newblock Dymo: Training-free diffusion model alignment with dynamic multi-objective scheduling.
\newblock In \emph{Proceedings of the Computer Vision and Pattern Recognition Conference}, pp.\  13220--13230, 2025.

\bibitem[Xue et~al.(2025)Xue, Wu, Gao, Kong, Zhu, Chen, Liu, Liu, Guo, Huang, et~al.]{xue2025dancegrpo}
Zeyue Xue, Jie Wu, Yu~Gao, Fangyuan Kong, Lingting Zhu, Mengzhao Chen, Zhiheng Liu, Wei Liu, Qiushan Guo, Weilin Huang, et~al.
\newblock Dancegrpo: Unleashing grpo on visual generation.
\newblock \emph{arXiv preprint arXiv:2505.07818}, 2025.

\bibitem[Yang et~al.(2024)Yang, Tao, Lyu, Ge, Chen, Shen, Zhu, and Li]{yang2024using}
Kai Yang, Jian Tao, Jiafei Lyu, Chunjiang Ge, Jiaxin Chen, Weihan Shen, Xiaolong Zhu, and Xiu Li.
\newblock Using human feedback to fine-tune diffusion models without any reward model.
\newblock In \emph{Proceedings of the IEEE/CVF Conference on Computer Vision and Pattern Recognition}, pp.\  8941--8951, 2024.

\bibitem[Zhang et~al.(2025)Zhang, Zheng, Wu, Zhang, Lin, Yu, Liu, Zhou, and Lin]{zhang2025lessons}
Zhenru Zhang, Chujie Zheng, Yangzhen Wu, Beichen Zhang, Runji Lin, Bowen Yu, Dayiheng Liu, Jingren Zhou, and Junyang Lin.
\newblock The lessons of developing process reward models in mathematical reasoning.
\newblock \emph{arXiv preprint arXiv:2501.07301}, 2025.

\end{thebibliography}
\bibliographystyle{arxiv}

\newpage
\appendix
\section{Appendix}
\subsection{Policy Gradient-Based Theoretical Framework}
\label{App1}

To provide a deeper understanding of our approach, we now examine it from the policy gradient perspective. \textbf{Note that in the summation $\sum_{k=0}^{T-1}$, the index $k$ denotes the timestep, while in subsequent equations, $k$ represents the timestep value.} Simplifying Equation~\ref{eq5}, we obtain $\bm{x}_{k-1} \sim \mathcal{N}(\mu_\theta(\bm{x}_k, k), \sigma_k^2\Delta k \bm{I})$, where:
\begin{equation}
\mu_\theta(\bm{x}_k, k) = \bm{x}_k + \left[\bm{v}_\theta(\bm{x}_k, k) + \frac{\sigma_k^2}{2k}\left(\bm{x}_k + (1-k)\bm{v}_\theta(\bm{x}_k, k)\right)\right]\Delta k
\end{equation}

Starting from the policy gradient formulation in Equation~\ref{eq8}, we have:
\begin{equation}
\nabla_\theta \mathcal{J}(\theta) = \sum_{k=0}^{T-1} \mathbb{E}_{\bm{x}_T \sim \mathcal{N}(0,\bm{I}), \bm{\epsilon} \sim \mathcal{N}(0,\bm{I})}[\nabla_\theta \log p_\theta(\bm{x}_{k-1}|\bm{x}_k) \hat{A}_k]
\end{equation}

Substituting $\bm{x}_{k-1}$ in the log-probability:
\begin{align}
\nabla_\theta \mathcal{J}(\theta) &= \sum_{k=0}^{T-1} \mathbb{E}_{\bm{x}_T \sim \mathcal{N}(0,\bm{I}), \bm{\epsilon} \sim \mathcal{N}(0,\bm{I})}\left[\nabla_\theta \log \exp\left(-\frac{\|\bm{x}_{k-1} - \mu_\theta(\bm{x}_k,k)\|^2}{2\sigma_k^2\Delta k}\right) \hat{A}_k\right] \\
&= \sum_{k=0}^{T-1} \mathbb{E}_{\bm{x}_T \sim \mathcal{N}(0,\bm{I}), \bm{\epsilon} \sim \mathcal{N}(0,\bm{I})}\left[\nabla_\theta \left(-\frac{\|\bm{x}_{k-1} - \mu_\theta(\bm{x}_k,k)\|^2}{2\sigma_k^2\Delta k}\right) \hat{A}_k\right]
\end{align}

Taking the gradient with respect to $\theta$:
\begin{equation}
\nabla_\theta \mathcal{J}(\theta) = \sum_{k=0}^{T-1} \mathbb{E}_{\bm{x}_T \sim \mathcal{N}(0,\bm{I}), \bm{\epsilon} \sim \mathcal{N}(0,\bm{I})}\left[\frac{\bm{x}_{k-1} - \mu_\theta(\bm{x}_k,k)}{\sigma_k^2\Delta k} \cdot \nabla_\theta \mu_\theta(\bm{x}_k,k) \hat{A}_k\right]
\end{equation}

Since $\bm{x}_{k-1} = \mu_\theta(\bm{x}_k,k) + \sigma_k\sqrt{\Delta k} \cdot \bm{\epsilon}$ where $\bm{\epsilon} \sim \mathcal{N}(0,\bm{I})$:
\begin{equation}
\nabla_\theta \mathcal{J}(\theta) = \sum_{k=0}^{T-1} \mathbb{E}_{\bm{x}_T \sim \mathcal{N}(0,\bm{I}), \bm{\epsilon} \sim \mathcal{N}(0,\bm{I})}\left[\frac{\bm{\epsilon}}{\sigma_k\sqrt{\Delta k}} \cdot \nabla_\theta \mu_\theta(\bm{x}_k,k) \hat{A}_k\right]
\end{equation}

Expanding $\nabla_\theta \mu_\theta(\bm{x}_k,k)$:
\begin{align}
\nabla_\theta \mu_\theta(\bm{x}_k,k) &= \nabla_\theta \left[\bm{x}_k + \left(\bm{v}_\theta(\bm{x}_k,k) + \frac{\sigma_k^2}{2k}\left(\bm{x}_k + (1-k)\bm{v}_\theta(\bm{x}_k,k)\right)\right)\Delta k\right] \\
&= \nabla_\theta \left[\left(\bm{v}_\theta(\bm{x}_k,k) + \frac{\sigma_k^2(1-k)}{2k}\bm{v}_\theta(\bm{x}_k,k)\right)\Delta k\right] \\
&= \left(1 + \frac{\sigma_k^2(1-k)}{2k}\right)\Delta k \cdot \nabla_\theta \bm{v}_\theta(\bm{x}_k,k)
\end{align}

Substituting back:
\begin{align}
\nabla_\theta \mathcal{J}(\theta) &= \sum_{k=0}^{T-1} \mathbb{E}_{\bm{x}_T \sim \mathcal{N}(0,\bm{I}), \bm{\epsilon} \sim \mathcal{N}(0,\bm{I})}\left[\frac{\bm{\epsilon}}{\sigma_k\sqrt{\Delta k}} \cdot \left(1 + \frac{\sigma_k^2(1-k)}{2k}\right)\Delta k \cdot \nabla_\theta \bm{v}_\theta(\bm{x}_k,k) \hat{A}_k\right] \\
&= \sum_{k=0}^{T-1} \mathbb{E}_{\bm{x}_T \sim \mathcal{N}(0,\bm{I}), \bm{\epsilon} \sim \mathcal{N}(0,\bm{I})}\left[\left(\frac{\sqrt{\Delta k}}{\sigma_k} + \frac{\sigma_k\sqrt{\Delta k}(1-k)}{2k}\right) \cdot \bm{\epsilon} \cdot \nabla_\theta \bm{v}_\theta(\bm{x}_k,k) \hat{A}_k\right]
\end{align}

With $\sigma_k = a\sqrt{\frac{k}{1-k}}$, we get:
\begin{align}
\frac{\sqrt{\Delta k}}{\sigma_k} &= \frac{\sqrt{\Delta k}}{a\sqrt{\frac{k}{1-k}}} = \frac{1}{a}\sqrt{\frac{\Delta k(1-k)}{k}} \\
\frac{\sigma_k\sqrt{\Delta k}(1-k)}{2k} &= \frac{a\sqrt{\frac{k}{1-k}}\sqrt{\Delta k}(1-k)}{2k} = \frac{a}{2}\sqrt{\frac{\Delta k(1-k)}{k}}
\end{align}

Therefore:
\begin{equation}
\nabla_\theta \mathcal{J}(\theta) = \sum_{k=0}^{T-1} \mathbb{E}_{\bm{x}_T \sim \mathcal{N}(0,\bm{I}), \bm{\epsilon} \sim \mathcal{N}(0,\bm{I})}\left[\left(\frac{1}{a} + \frac{a}{2}\right)\underbrace{\sqrt{\frac{\Delta k(1-k)}{k}}}_\text{Scale Term} \cdot \bm{\epsilon} \cdot \nabla_\theta \bm{v}_\theta(\bm{x}_k,k) \hat{A}_k\right]
\label{eq22}
\end{equation}

This reveals that the natural gradient coefficient is proportional to $\sqrt{\frac{1-k}{k}}\sqrt{\Delta k}$, which captures the intrinsic exploration potential at timestep $k$. After reweighting, we have the following derivation (don't consider $a$ and norm coefficient):
\begin{equation}
\nabla_\theta \mathcal{J}(\theta) = \sum_{k=0}^{T-1} \mathbb{E}_{\bm{x}_T \sim \mathcal{N}(0,\bm{I}), \bm{\epsilon} \sim \mathcal{N}(0,\bm{I})}\left[\left(\frac{1}{a} + \frac{a}{2}\right)\underbrace{\Delta k}_\text{Scale Term} \cdot \bm{\epsilon} \cdot \nabla_\theta \bm{v}_\theta(\bm{x}_k,k) \hat{A}_k\right]
\end{equation}

Consider $\mathbb{E}_{\bm{\epsilon}\sim\mathcal{N}(0,\bm{I})}[\bm{\epsilon} \hat{A}_k]$, suppose the final reward is a function of the small noise vector $\sigma_k\sqrt{\Delta k}\bm{\epsilon}$ applied at a certain step. When $\sigma_k\sqrt{\Delta k}\bm{\epsilon}$ is small (and drawn from a zero-mean Gaussian), we can approximate the reward using a first-order Taylor expansion:
\begin{equation}
    R_k(\sigma_k\sqrt{\Delta k}\bm{\epsilon}) \approx R_k(0) + \sigma_k\sqrt{\Delta k}\bm{\epsilon}^T\nabla_{\sigma_k\sqrt{\Delta k}\bm{\epsilon}} R_k|_{\sigma_k\sqrt{\Delta k}\bm{\epsilon}=0}
\end{equation}

Since $\hat{A}_k$ is normalized version of $R_k$, the mean and std are as follows (let $\nabla_{\sigma_k\sqrt{\Delta k}\bm{\epsilon}} R_k|_{\sigma_k\sqrt{\Delta k}\bm{\epsilon}=0}=g_k$):
\begin{equation}
\begin{split}
    \text{mean} &= \mathbb{E}_{\sigma_k\sqrt{\Delta k}\bm{\epsilon}}[R_k(0) + \sigma_k\sqrt{\Delta k}\bm{\epsilon}^Tg_k] = R_k(0) \\
    \text{std} &= \sqrt{\mathbb{E}_{\sigma_k\sqrt{\Delta k}\bm{\epsilon}}[(R_k(\sigma_k\sqrt{\Delta k}\bm{\epsilon})-\text{mean})^2]} \\
    & = \sqrt{\mathbb{E}_{\sigma_k\sqrt{\Delta k}\bm{\epsilon}}[(\sigma_k\sqrt{\Delta k}\bm{\epsilon}^Tg_k)^2]}\\
    & = \sigma_k\sqrt{\Delta k}||g_k||
\end{split}
\end{equation}

Therefore:
\begin{equation}
\begin{split}
    &\hat{A}_k = \frac{R_k - \text{mean}}{\text{std}} 
 = \frac{\sigma_k\sqrt{\Delta k}\bm{\epsilon}^Tg_k}{\sigma_k\sqrt{\Delta k}||g_k||} \\
&\mathbb{E}_{\bm{\epsilon}}[\bm{\epsilon}\hat{A}_k]=\mathbb{E}_{\bm{\epsilon}}[\frac{\bm{\epsilon}\bm{\epsilon}^Tg_k}{||g_k||}] 
 = \frac{g_k}{||g_k||}
\end{split}
\label{eq28}
\end{equation}

Equation~\ref{eq28} indicates that the norm of $\mathbb{E}_{\bm{\epsilon}}[\bm{\epsilon}\hat{A}_k]$ is invariant among the timesteps.

\subsection{Experimental Setting Details}
\label{app2}

\noindent \textbf{Dataset.} We evaluate Compositional Image Generation on Geneval and Human Preference Alignment on PickScore. Meanwhile, we also test our method on HPDv2~\citep{wu2023human}.

\noindent \textbf{Reward Model.} In TempFlow-GRPO, we introduce four reward models, including clip-based PickScore, HPSv2~\citep{wu2023human}, VLM-based HPSv3~\citep{ma2025hpsv3} and framework Geneval.

\noindent \textbf{Human Preference Alignment.} Following Flow-GRPO, we evaluate our method on PickScore prompts using the PickScore reward function. PickScore is a CLIP-based scoring function that demonstrates superhuman performance in predicting human preferences for generated images. To ensure consistency with Flow-GRPO, we set the KL divergence regularization coefficient $\beta = 0.001$.

\noindent \textbf{Compositional Image Generation.} We adopt the GenEval framework following Flow-GRPO's experimental protocol. The training prompts are sourced from the Flow-GRPO dataset. GenEval provides an object-focused evaluation framework that assesses compositional image properties, including object co-occurrence, spatial positioning, object count, and color attributes. For fair comparison, we maintain the KL divergence regularization coefficient at $\beta = 0.004$.

\noindent \textbf{HPSv3 Evaluation on FLUX.1-dev.} For experiments on the FLUX.1-dev model, we configure the noise level at 0.9 for both Flow-GRPO and our TempFlow-GRPO method. The KL divergence weight is set to 0.004. During training, we employ 10 sampling steps, while evaluation uses 50 steps to ensure high-quality generation. The batch configuration consists of a group size of 24 (arranged as $4 \times 6$) with 48 groups in total.

\noindent \textbf{Comparison with DanceGRPO.} DanceGRPO and Flow-GRPO represent concurrent developments in this research area. The SDE sampling methodology employed in DanceGRPO corresponds to Equation 12 in the Flow-GRPO formulation. However, the theoretical derivation in DanceGRPO terminates at its Equation 7 without developing an analogous final formulation. To comprehensively validate our method's effectiveness, we conduct a direct comparison with DanceGRPO. For equitable comparison, we train TempFlow-GRPO for 300 iterations on the HPDv2 dataset using the HPSv2 reward function.

\subsection{Experiments on FLUX.1-dev}
\label{app3}

\begin{figure}[h]
    \centering
    \begin{subfigure}[b]{0.48\textwidth}
        \centering
        \includegraphics[width=\textwidth]{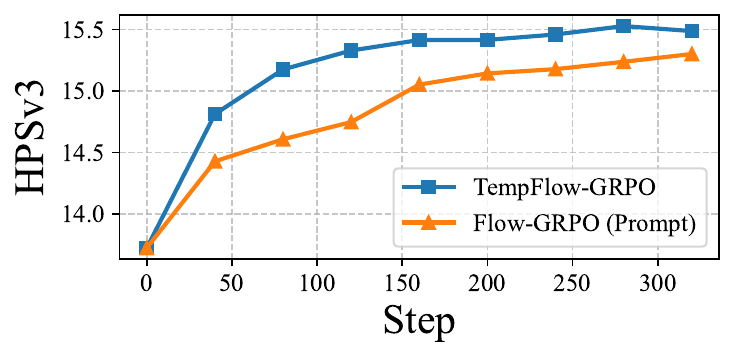}
        \label{app_fig2_1}
    \end{subfigure}
    \hfill
    \begin{subfigure}[b]{0.48\textwidth}
        \centering
        \includegraphics[width=\textwidth]{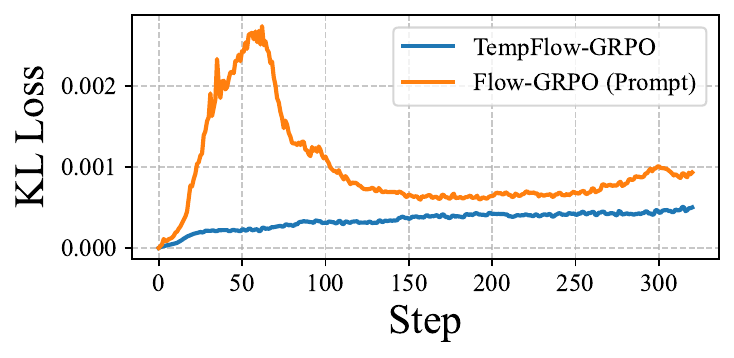}
        \label{app_fig2_2}
    \end{subfigure}
    \vspace{-0.5cm}
    \caption{The performance and KL loss (smoothed) of HPSv3 on HPDv2 dataset with FLUX.1-dev as base model.}
    \label{app_fig2}
\end{figure}

We further evaluate the performance of our method against Flow-GRPO on FLUX.1-dev, utilizing the HPDv2 dataset with HPSv3 as the reward model. As illustrated in Figure~\ref{app_fig2}, our method demonstrates substantial improvements over the baseline. After 300 training steps, our approach achieves a performance gain of approximately 0.3 compared to Flow-GRPO. More significantly, our method exhibits superior training efficiency: it reaches the performance level that Flow-GRPO achieves at 300 steps in merely 80 steps, representing a 3.75$\times$ speedup in convergence. Additionally, the results indicate that our method maintains lower and more stable KL divergence values throughout the optimization process, suggesting improved training dynamics and better preservation of the original model's distribution compared to Flow-GRPO.

\subsection{Comparsion with DanceGRPO}
\label{app4}

As illustrated in Figure~\ref{app_fig3} (Left), we present the evaluation curves of our method on the HPDv2 dataset. The results demonstrate clear superiority of our approach over the baseline. While DanceGRPO achieves a final performance score of 37.2 after 300 iterations, our method attains a higher score of 38.5 at 280 iterations, representing a 1.3\% improvement over the DanceGRPO. More significantly, our approach exhibits substantially enhanced training efficiency: it matches DanceGRPO's final performance (achieved at 300 iterations) in merely 150 iterations, yielding a 2$\times$ speedup in convergence. Notably, this superior performance is achieved without any specialized hyperparameter tuning, as we directly adopt the configuration settings from Flow-GRPO, further validating the robustness and generalizability of our method.

\subsection{Experiments on SD3.5-M with 1024$\times$1024 resolution}
\label{app5}
\begin{figure}[t]
    \centering
    \begin{subfigure}[b]{0.48\textwidth}
        \centering
        \includegraphics[width=\textwidth]{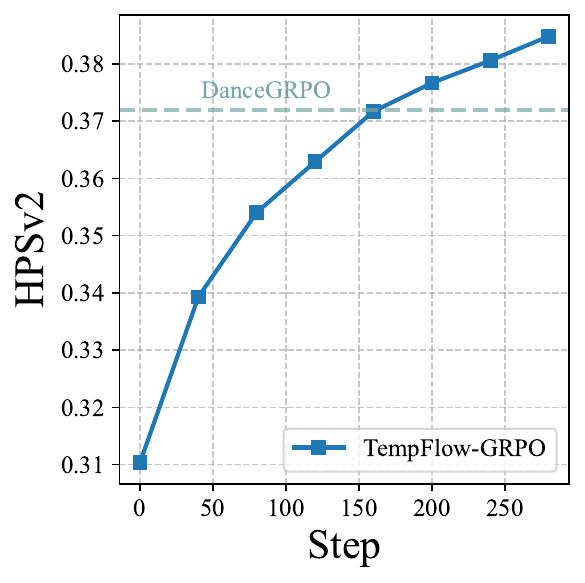}
        \label{app_fig3_1}
    \end{subfigure}
    \hfill
    \begin{subfigure}[b]{0.48\textwidth}
        \centering
        \includegraphics[width=\textwidth]{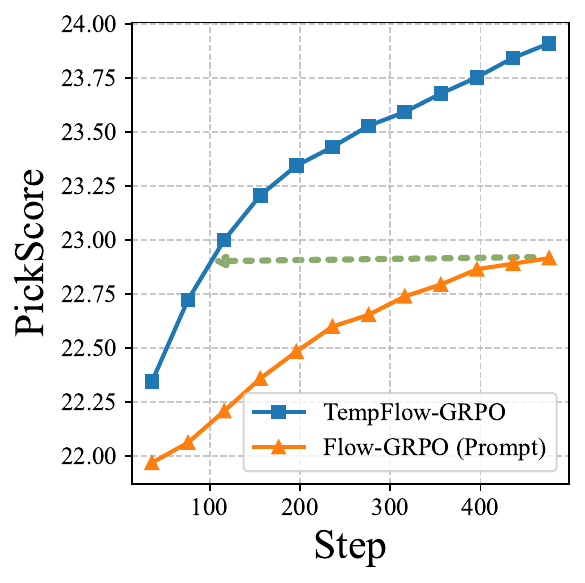}
        \label{app_fig3_2}
    \end{subfigure}
    \vspace{-0.5cm}
    \caption{(Left) Performance of TempFlow-GRPO on HPDv2 with HPSv2 as reward compared DanceGRPO. (Right) Comparsion on PickScore
    benchmark (PickScore reward, SD3.5-M 1024).}
    \label{app_fig3}
\end{figure}

We investigate the effectiveness of our approach at higher image resolutions, employing PickScore as the reward model for evaluation. As illustrated in Figure~\ref{app_fig3} (Right), TempFlow-GRPO achieves a 1.0\% improvement in PickScore after 450 training steps compared to the Flow-GRPO. More notably, our method demonstrates exceptional training efficiency, requiring only approximately 100 steps to match the performance that Flow-GRPO (Prompt) achieves after 450 steps—a 4.5$\times$ speedup. This result further validates the efficiency and effectiveness of our proposed method across varying image resolutions.
\subsection{Training Hours}
\begin{figure}[t]
    \centering
    \begin{subfigure}[b]{0.48\textwidth}
        \begin{subfigure}[b]{\textwidth}
            \centering
            \includegraphics[width=\textwidth]{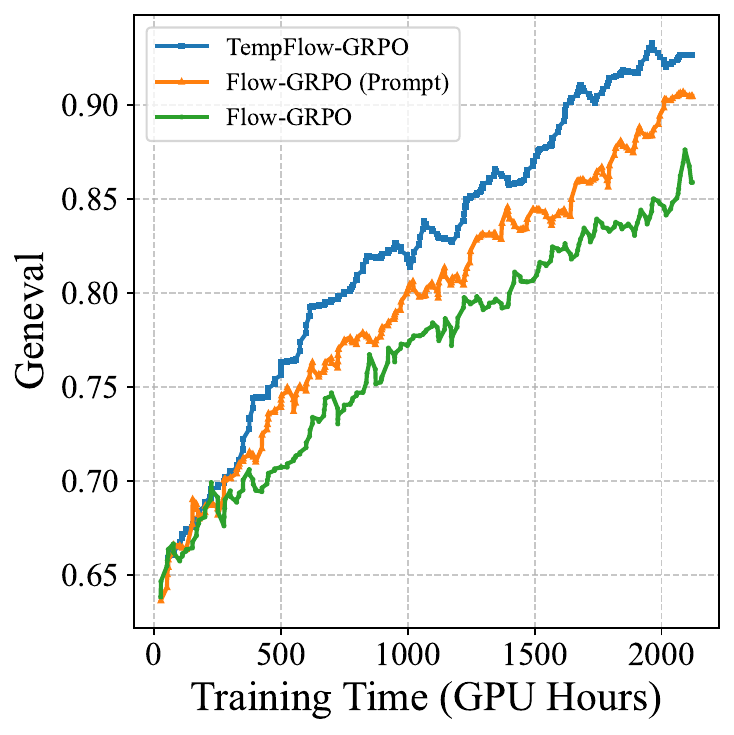}
            \label{app_fig4_1}
            \vspace{-0.5cm}
            \caption{Training time comparison on GenEval benchmark (Geneval reward, SD3.5-M 512).}
        \end{subfigure}
        \begin{subfigure}[b]{\textwidth}
            \centering
            \includegraphics[width=\textwidth]{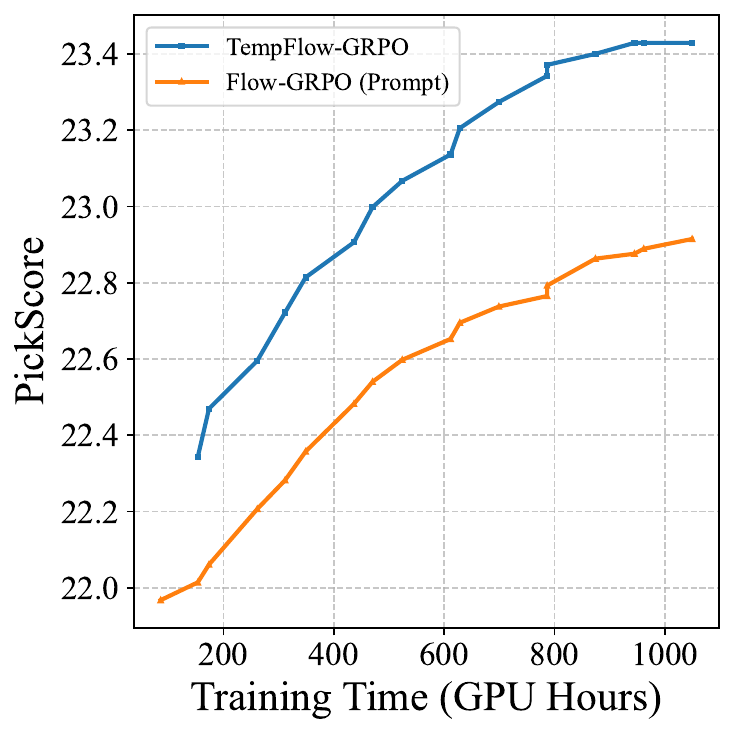}
            \label{app_fig4_2}
            \vspace{-0.5cm}
            \caption{Training time comparison on PickScore prompts (PickScore reward, SD3.5-M 1024).}
        \end{subfigure}
    \end{subfigure}
    \hfill
    \begin{subfigure}[b]{0.48\textwidth}
        \begin{subfigure}[b]{\textwidth}
            \centering
            \includegraphics[width=\textwidth]{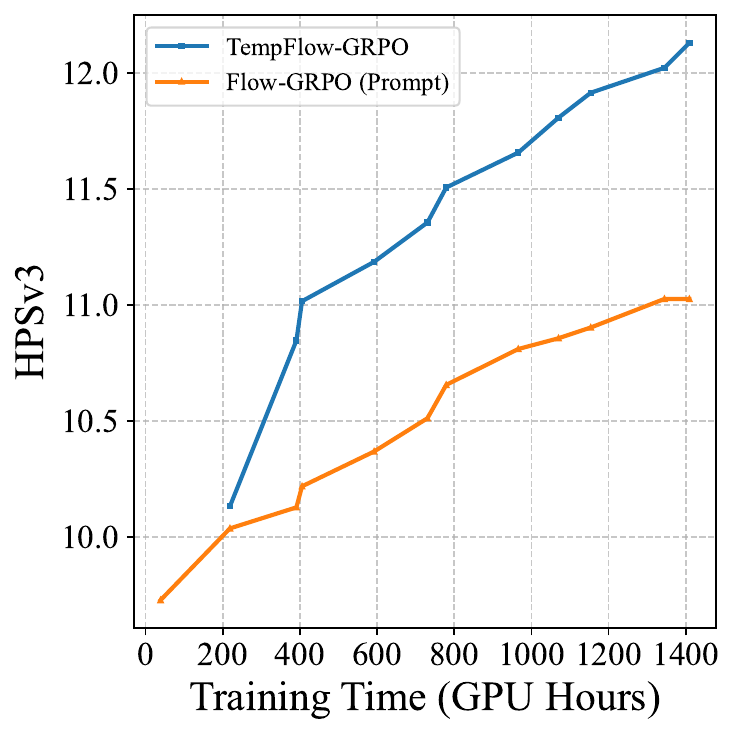}
            \label{app_fig4_3}
            \vspace{-0.5cm}
            \caption{Training time comparison on PickScore prompts (HPSv3 as reward, FLUX.1-dev 1024).}
        \end{subfigure}
        \begin{subfigure}[b]{\textwidth}
            \centering
            \includegraphics[width=\textwidth]{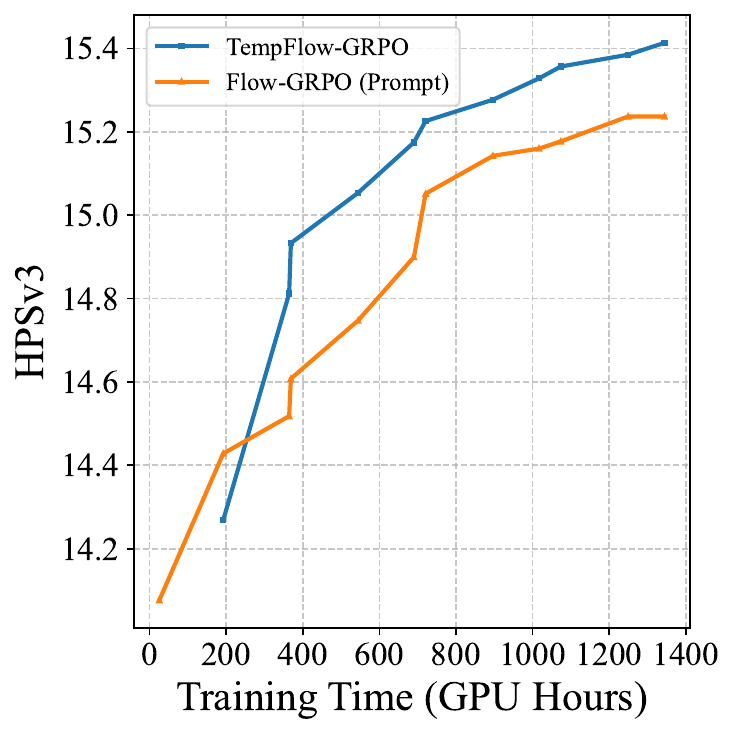}
            \label{app_fig4_4}
            \vspace{-0.5cm}
            \caption{Training time comparison on HPDv2 prompts (HPSv3 reward, FLUX.1-dev 1024).}
        \end{subfigure}
    \end{subfigure}
    \caption{Comprehensive comparison of training time across multiple benchmarks and resolutions. TempFlow-GRPO consistently achieves superior performance with reduced training time compared to Flow-GRPO baseline.}
    \label{app_fig4}
\end{figure}

\begin{figure}[!t]
    \centering
    \begin{subfigure}[b]{0.3\textwidth}
        \centering
        \includegraphics[width=\textwidth]{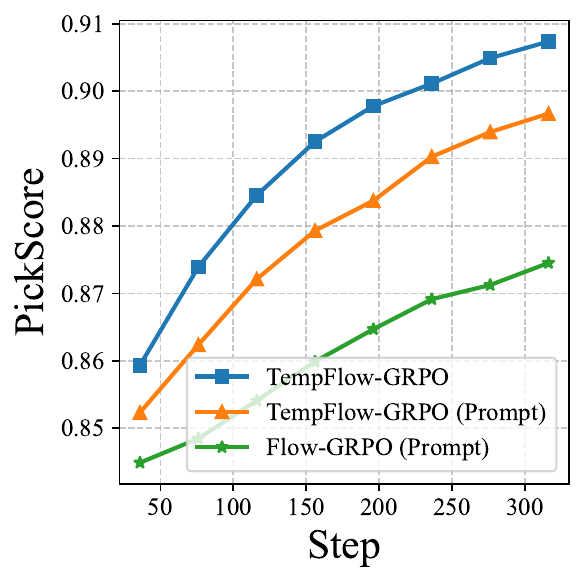}
        \caption{Group strategy analysis on PickScore prompts (PickScore as reward, SD3.5-M 1024).}
        \label{app_fig5_1}
    \end{subfigure}
    \begin{subfigure}[b]{0.3\textwidth}
        \centering
        \includegraphics[width=\textwidth]{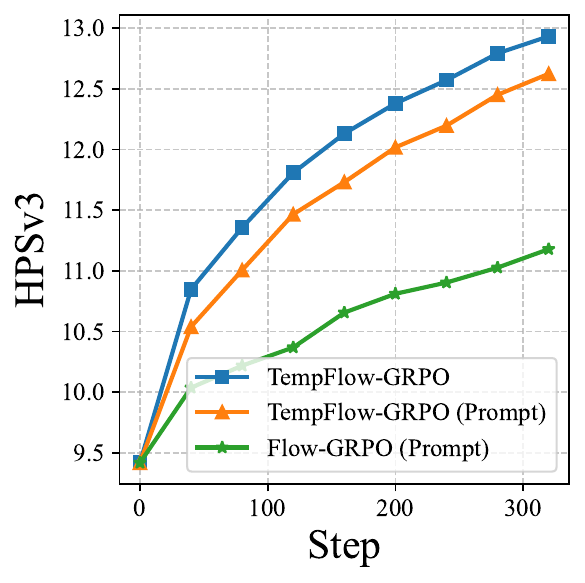}
        \caption{Group strategy analysis on PickScore prompts (HPSv3 as reward, FLUX.1-dev 1024).}
        \label{app_fig5_2}
    \end{subfigure}
    \begin{subfigure}[b]{0.3\textwidth}
        \centering
        \includegraphics[width=\textwidth]{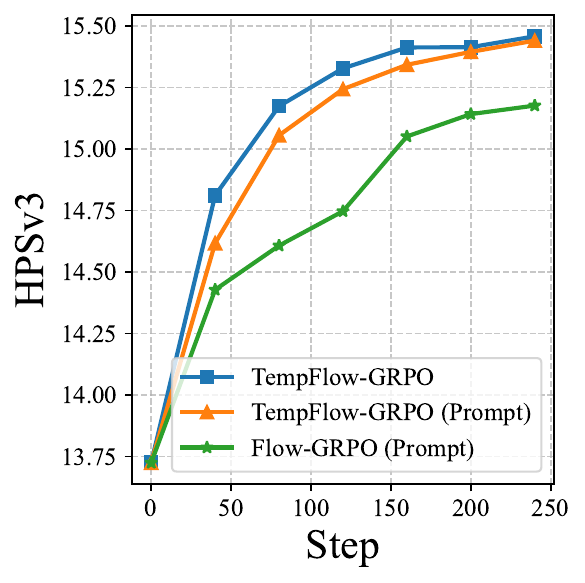}
        \caption{Group strategy analysis  on HPDv2 prompts (HPSv3 as reward, FLUX.1-dev 1024).}
        \label{app_fig5_3}
    \end{subfigure}
    \caption{Comprehensive analysis of group strategy impact across different models and benchmarks, demonstrating consistent performance improvements with our proposed method and seed group approach.}
    \label{app_fig5}
\end{figure}

A discussion of the computational cost of our method is warranted. Due to the $K$-branch exploration at each timestep, our sampling process incurs higher computational overhead compared to Flow-GRPO. Specifically, for $K=10$, the average number of branches is approximately $(9+\ldots+1)/10 = 4.5$ times that of Flow-GRPO. However, the training time per iteration remains identical to Flow-GRPO. Despite this increased per-sample computational cost, our method demonstrates substantially superior overall training efficiency. As illustrated in Figure~\ref{fig2} (bottom left, main paper) and Figure~\ref{app_fig4}, where we plot performance against wall-clock training time, our method exhibits faster convergence across all evaluated benchmarks. This efficiency gain becomes particularly pronounced at higher resolutions, where our approach achieves Flow-GRPO's final performance using only 33\% to 50\% of the total training time.

Furthermore, our approach not only converges faster but also achieves superior final performance. The PickScore experiments presented in the main paper reveal that even after extended training of 1200 steps, our method continues to outperform Flow-GRPO, providing compelling evidence that our approach possesses a higher performance ceiling.

\subsection{Additional Details on Group Strategy}

In Section~\ref{sec_4_3} of the main paper, we analyze the impact of the group strategy on performance. Figure~\ref{fig5} and Figure~\ref{app_fig5} demonstrate that TempFlow-GRPO achieves substantial improvements even when employing an identical group strategy to the baseline, indicating the inherent advantages of our proposed core modules. To systematically disentangle the contributions of the group strategy from our two fundamental components—trajectory branching and noise reweighting—we conduct controlled ablation studies. In these experiments, we maintain a constant group strategy configuration to ensure fair comparison. The results unequivocally demonstrate that our core modules independently yield significant performance gains. These components not only enable the model to match Flow-GRPO's multi-step performance in substantially fewer iterations but also elevate the final performance ceiling, validating the effectiveness of our theoretical framework.

\subsection{KL Divergence Analysis}
\begin{figure}[t]
    \centering
    \begin{subfigure}[b]{0.45\textwidth}
        \centering
        \includegraphics[width=\textwidth]{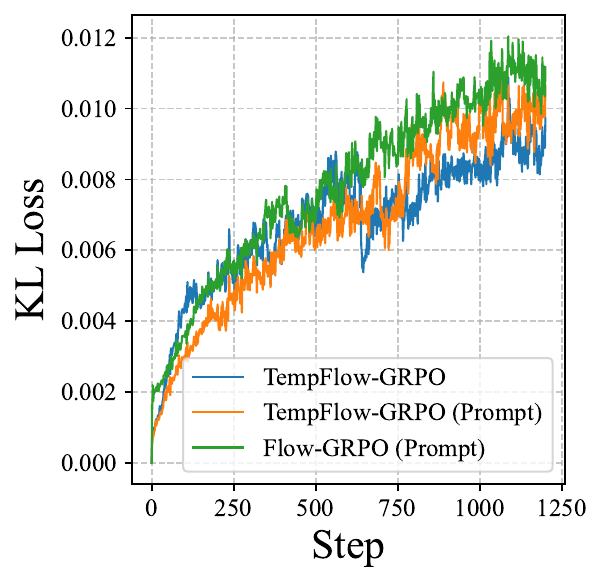}
        \caption{KL Loss on PickScore (SD3.5-M 512).}
        \label{app_fig6_1}
    \end{subfigure}
    \begin{subfigure}[b]{0.45\textwidth}
        \centering
        \includegraphics[width=\textwidth]{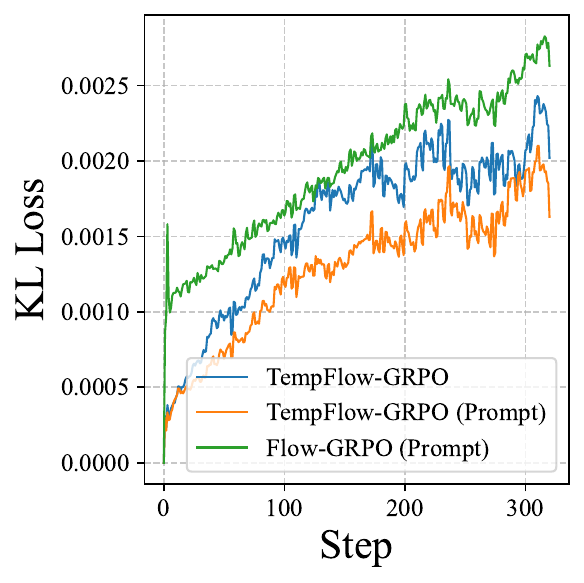}
        \caption{KL Loss on PickScore (SD3.5-M 1024).}
        \label{app_fig6_2}
    \end{subfigure}
    \caption{Comprehensive analysis of KL divergence dynamics during training across different resolutions. TempFlow-GRPO consistently maintains lower KL divergence values compared to Flow-GRPO, indicating better preservation of the original model distribution while achieving superior preference alignment.}
    \label{app_fig6}
\end{figure}

To further substantiate the training stability of our method, we provide a comprehensive analysis of the KL divergence dynamics during optimization. The results, presented in Figure~\ref{app_fig6}, clearly demonstrate that our method consistently maintains significantly lower KL divergence values throughout the training process compared to Flow-GRPO. This reduced divergence indicates better preservation of the original model distribution while still achieving effective preference alignment.

Furthermore, we analyze the impact of different grouping strategies on KL divergence. While adopting the ``seed group" strategy results in a marginal increase in KL divergence compared to our default configuration, the values remain substantially below those of the Flow-GRPO baseline across all training iterations. This observation provides strong evidence that our core modules—trajectory branching and noise reweighting—inherently enhance training stability and robustly maintain distributional proximity, regardless of the specific grouping strategy employed. The consistent stability across different configurations underscores the fundamental advantages of our theoretical framework in balancing exploration and exploitation during the optimization process.

\subsection{Multi-Reward Optimization}
\begin{table}[htbp]   
    \setlength{\tabcolsep}{10mm}
    \centering
    \caption{Performance comparison for multi-reward optimization on PickScore prompts, demonstrating simultaneous improvement across multiple objectives.}
     \begin{tabular}{l|cc}
        \toprule
         & HPSv3 $\uparrow$ & PickScore $\uparrow$ \\
         \midrule
         FLUX.1-dev & 9.42 & 22.34 \\
         \midrule
         Flow-GRPO (Prompt) & 10.41 (+0.99) & 22.44 (+0.1)  \\
         TempFlow-GRPO & 12.23 (+1.81) & 22.52 (+0.18)\\
        \bottomrule
    \end{tabular}
    \label{app_tab2}
\end{table}

We evaluate the performance of our method in multi-reward optimization scenarios. For this experiment, we employ FLUX.1-dev as the base model and train at a high resolution of 1024$\times$1024. We simultaneously optimize for two reward functions—HPSv3 and PickScore—with a weight ratio of 1:0.26. After 120 training steps, the results presented in Table~\ref{app_tab2} demonstrate our method's robust capability for multi-objective optimization. Compared to Flow-GRPO, our approach achieves a substantial improvement of 0.82 on the HPSv3 metric while simultaneously maintaining consistent performance gains on PickScore. These results validate that our TempFlow-GRPO effectively balances multiple objectives without sacrificing performance on individual metrics, highlighting the versatility of our approach in complex preference alignment scenarios where multiple criteria must be optimized concurrently.

\subsection{Visualization}
We present comprehensive qualitative comparisons to demonstrate the superior visual quality achieved by our TempFlow-GRPO method. The visualization results consistently show that our approach generates images with enhanced fidelity, better adherence to complex prompts, and fewer visual artifacts compared to baseline methods.

\noindent \textbf{Prompts in Figure~1. } The prompts in Figure~1 are as follows:
\begin{tcolorbox}[colback=white]
1. 16-year-old teenager wearing a white bear-ear hat with a smirk on their face. \\
2. photo of well done salmon dinner, 8K, Global Illumination, Ray Tracing Reflections \\
3. A lemon with a McDonald's hat. \\
4. An image of an emo with dark brown hair in a messy pixie cut, large entirely-black eyes, wearing black clothing. \\
5. The image is a mixed media collage with broken glass and torn paper elements, featuring intricate oil details and a canvas texture, in a contemporary art style. \\
6. An epic deep space photograph. Gigantic, monolithic letters forming the word 'TempFlow-' hang silently in the void, their surfaces like ancient, cracked obsidian reflecting distant starfields. Far below, the letters 'GRPO' are formed by a vast, tranquil nebula glowing with soft, ethereal light, like a cosmic ocean. The sense of scale is immense and humbling. Perfect, case-sensitive lettering. Moody, atmospheric, photorealistic, cinematic wide-angle shot, 4K UHD.\\
7. Kiwi fruit, mint leaves, ice cubes, background yellow, splashing water, soft box, back light, creative food photography, Art by Alberto Seveso,
\\
8. Claymation of Futurama characters. \\
9. A group of four friends commemorating a ski trip in the snow. \\
10. an empty bench next to a busy street. \\
11. a 12 year old girl and her pet raccoon \\
\end{tcolorbox}

\begin{figure*}[h]
    \centering
    \includegraphics[width=0.9\linewidth]{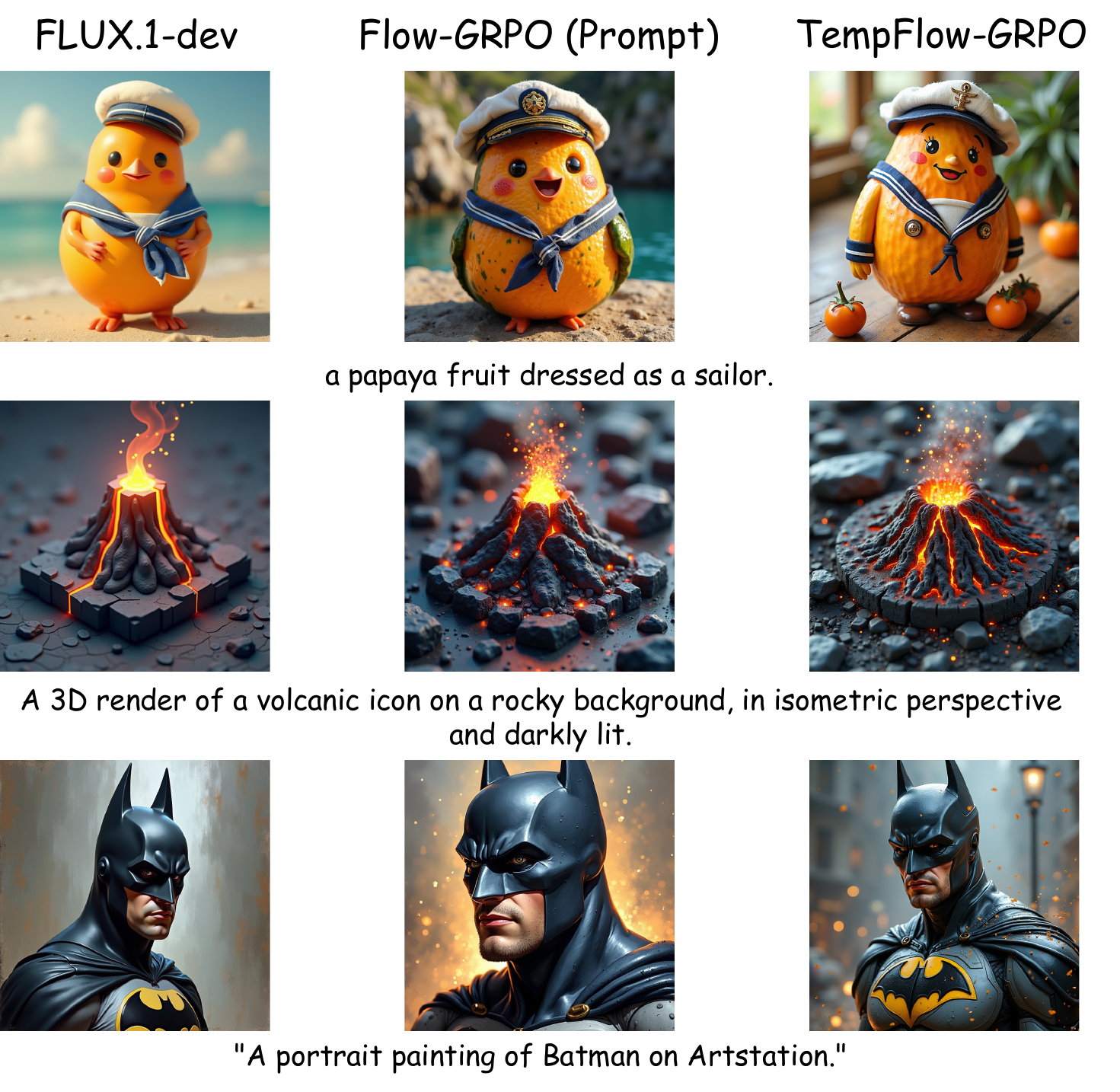}
    \caption{Qualitative comparison between FLUX.1-dev, Flow-GRPO (Prompt) and TempFlow-GRPO with HPSv3 rewards on HPDv2 prompts.}
    \includegraphics[width=0.9\linewidth]{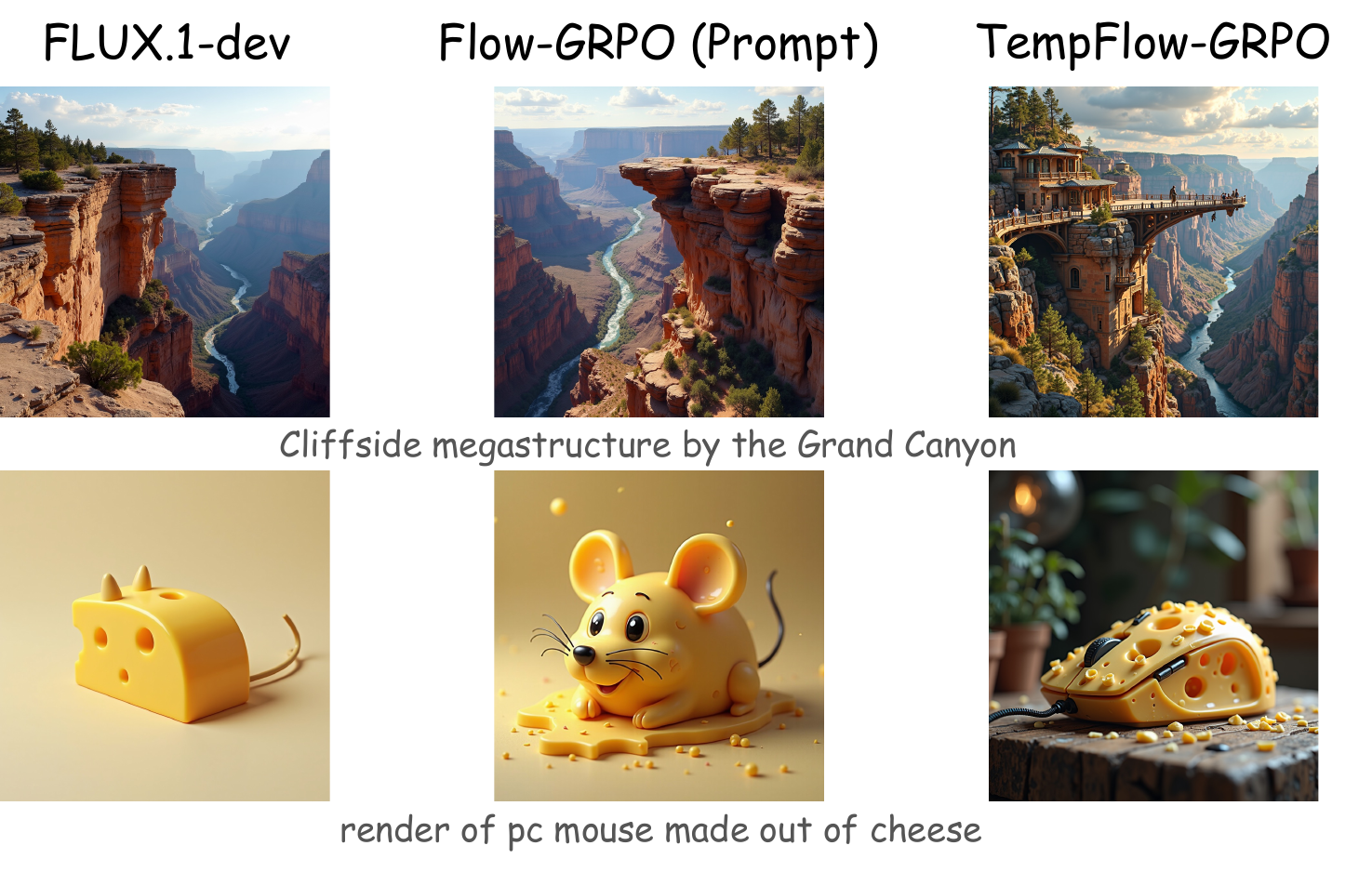}
    \caption{Qualitative comparison between FLUX.1-dev, Flow-GRPO (Prompt) and TempFlow-GRPO with multi rewards on PickScore prompts.}
\end{figure*} 

\begin{figure*}[t]
    \centering
    \includegraphics[width=0.9\linewidth]{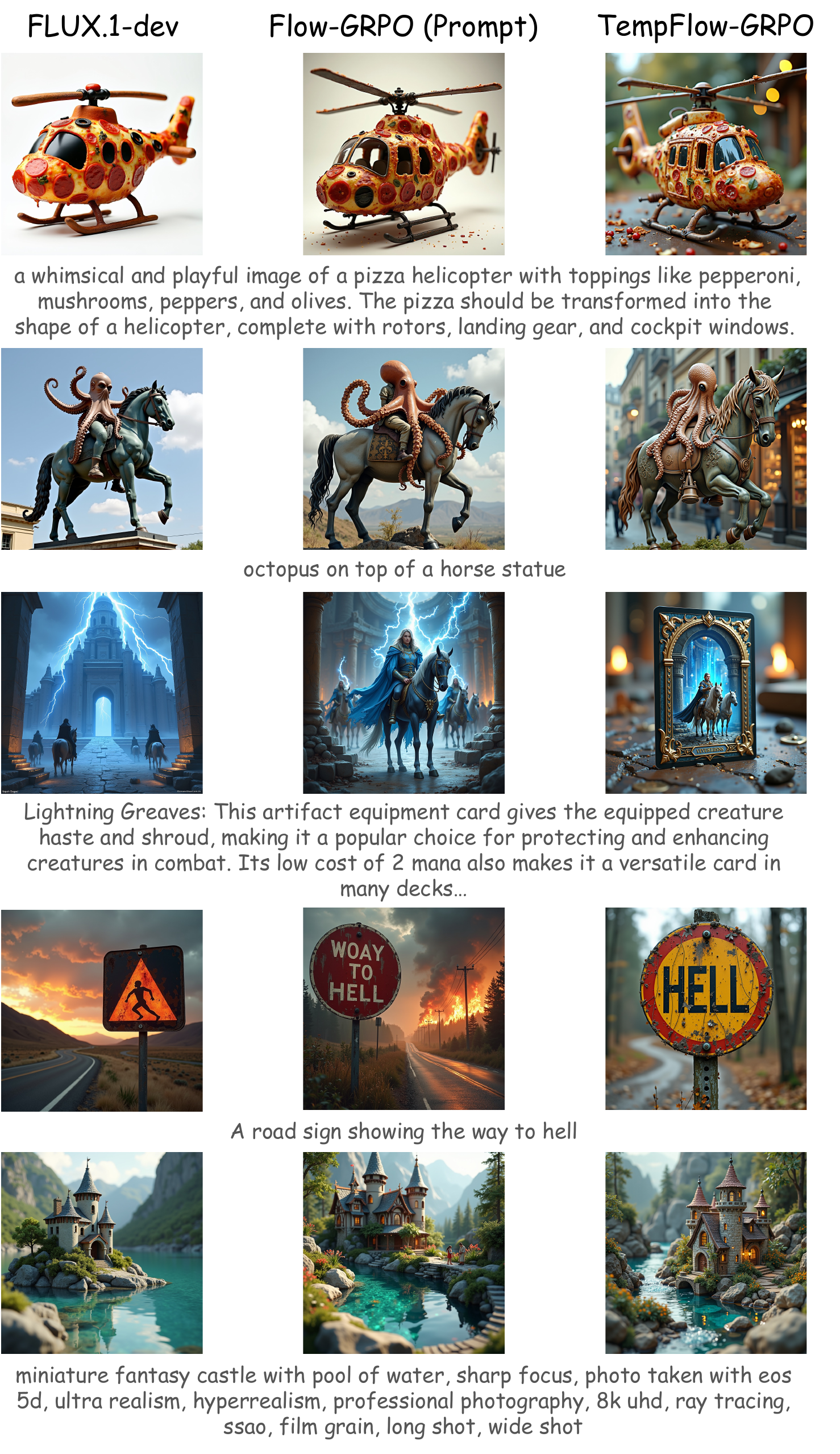}
    \caption{Qualitative comparison between FLUX.1-dev, Flow-GRPO (Prompt) and TempFlow-GRPO with multi rewards on PickScore prompts.}
\end{figure*} 
\begin{figure*}[t]
    \centering
    \includegraphics[width=0.9\linewidth]{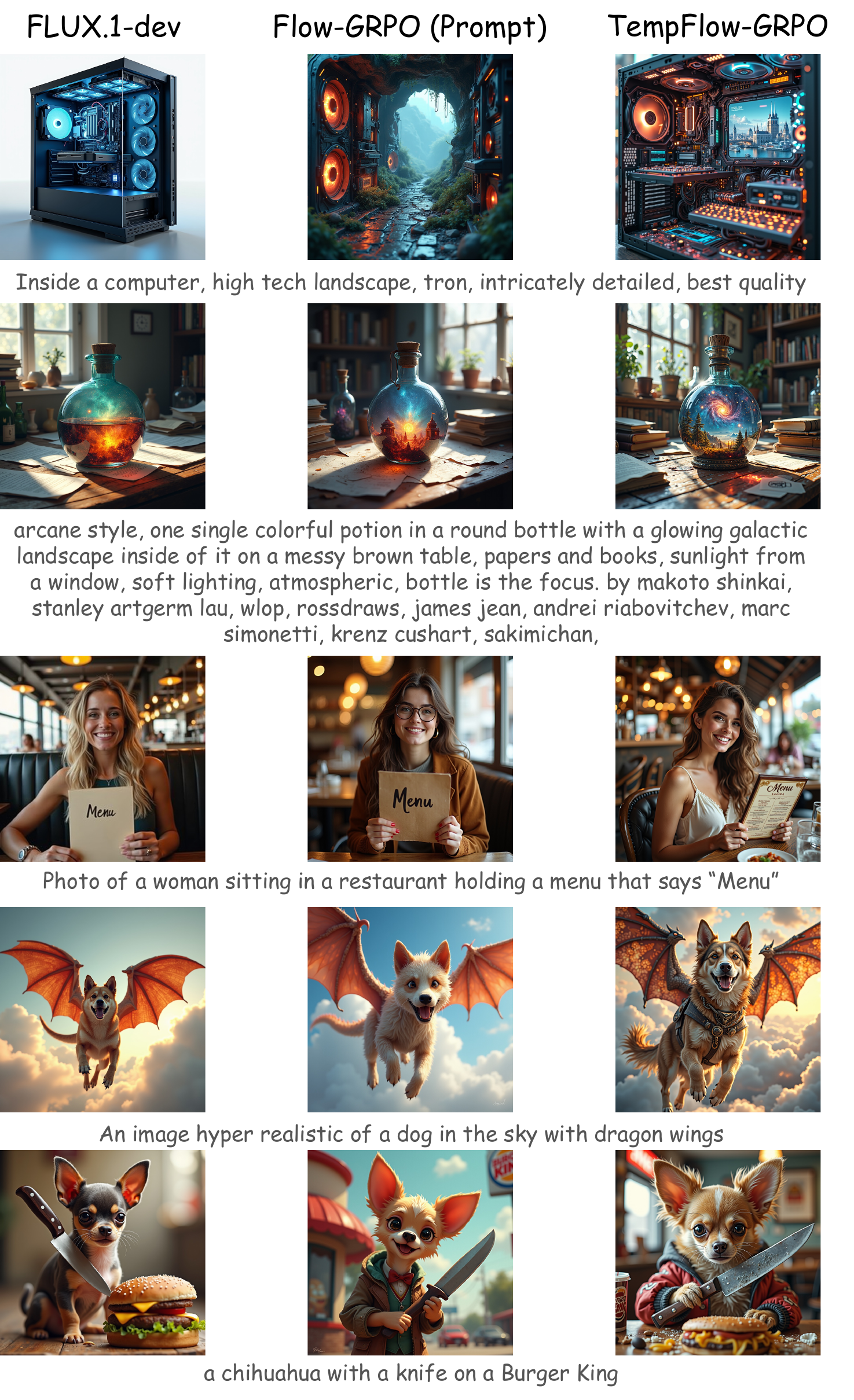}
    \caption{Qualitative comparison between FLUX.1-dev, Flow-GRPO (Prompt) and TempFlow-GRPO with multi rewards on PickScore prompts.}
\end{figure*} 
\begin{figure*}[t]
    \centering
    \includegraphics[width=0.9\linewidth]{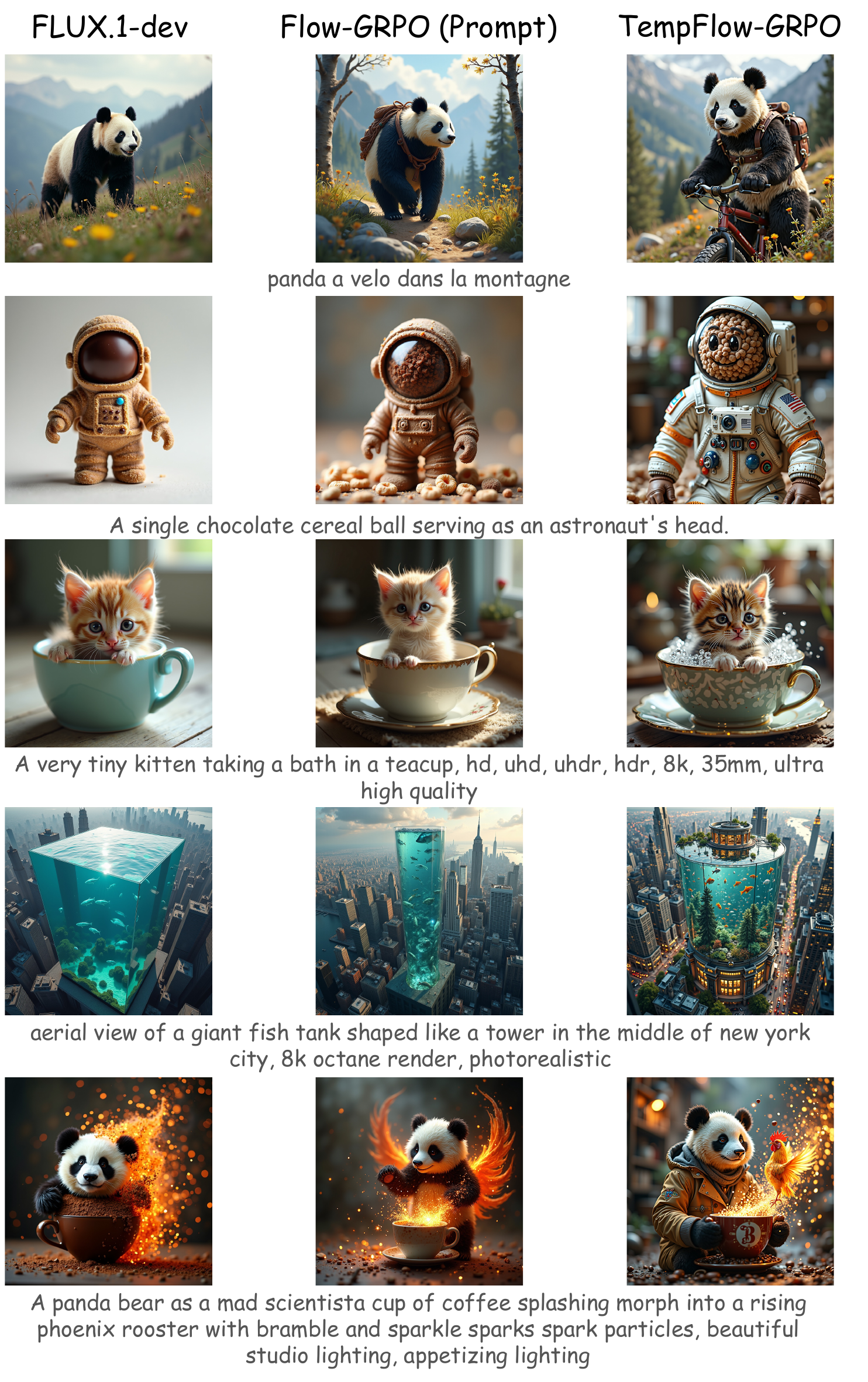}
    \caption{Qualitative comparison between FLUX.1-dev, Flow-GRPO (Prompt) and TempFlow-GRPO with multi rewards on PickScore prompts.}
\end{figure*} 

\end{document}